\documentclass[11pt]{article}

\usepackage{float}
\usepackage[final]{acl}
\usepackage{enumitem}
\usepackage{needspace}
\usepackage{dblfloatfix}
\usepackage{times}
\usepackage{latexsym}
\usepackage{algpseudocode}

\usepackage[T1]{fontenc}

\usepackage[utf8]{inputenc}

\usepackage{microtype}

\usepackage{inconsolata}

\usepackage{graphicx}
\usepackage{multirow, tabularx}
\usepackage{wrapfig}
\usepackage{booktabs}
\usepackage{hyperref}
\usepackage{url}
\usepackage{amsmath}
\usepackage{subcaption}

\usepackage{listings}
\newcommand\blfootnote[1]{%
  \begingroup
  \renewcommand\thefootnote{}\footnote{#1}%
  \addtocounter{footnote}{-1}%
  \endgroup
}

\title{RoboReason: Boosting Robotics-Relevant VLM Reasoning Through Scalable Data Generation}

\author{
\textbf{Sagnik Anupam\textsuperscript{*},} 
\textbf{Utkarsh Lal\textsuperscript{*}, }
\textbf{Aneesh Edara, }
\textbf{Botong Zhang, }
\textbf{Yao Tang, } \\
\textbf{Jiatao Gu, }
\textbf{Dinesh Jayaraman\textsuperscript{\textdagger}, }
\textbf{Osbert Bastani\textsuperscript{\textdagger}} \\
University of Pennsylvania
}

\begin{document}
\maketitle
\begin{abstract}
Vision-Language Models (VLMs) have demonstrated impressive performance when integrated into robot manipulation pipelines, but real-world evaluation of VLM capabilities on downstream robotics tasks is both difficult and expensive to scale. We propose that spatial and temporal ordering tasks using images from stereo and wrist cameras attached to real-world robots are particularly well-suited for evaluating visual reasoning capabilities, as they assess skills directly applicable to robots. We evaluate VLM reasoning abilities on three questions: ordering shuffled frames conditioned on a language description of the task, ordering objects depthwise based on distance from the camera, and ordering objects from left-to-right in pixel coordinates. The complexity of these tasks can also be modulated by changing the size of the set to sort. Our experiments show that several state-of-the-art vision-language models struggle on ordering tasks. To fix this, we propose a scalable autonomous training data generation pipeline, and show that both supervised finetuning and reinforcement-learning on the generated data can significantly improve VLM reasoning performance both on spatiotemporal ordering tasks as well as downstream robotics and visual reasoning tasks.
\end{abstract}

\section{Introduction}

\begin{figure*}[htbp]
\centering
\includegraphics[width=0.6\linewidth]{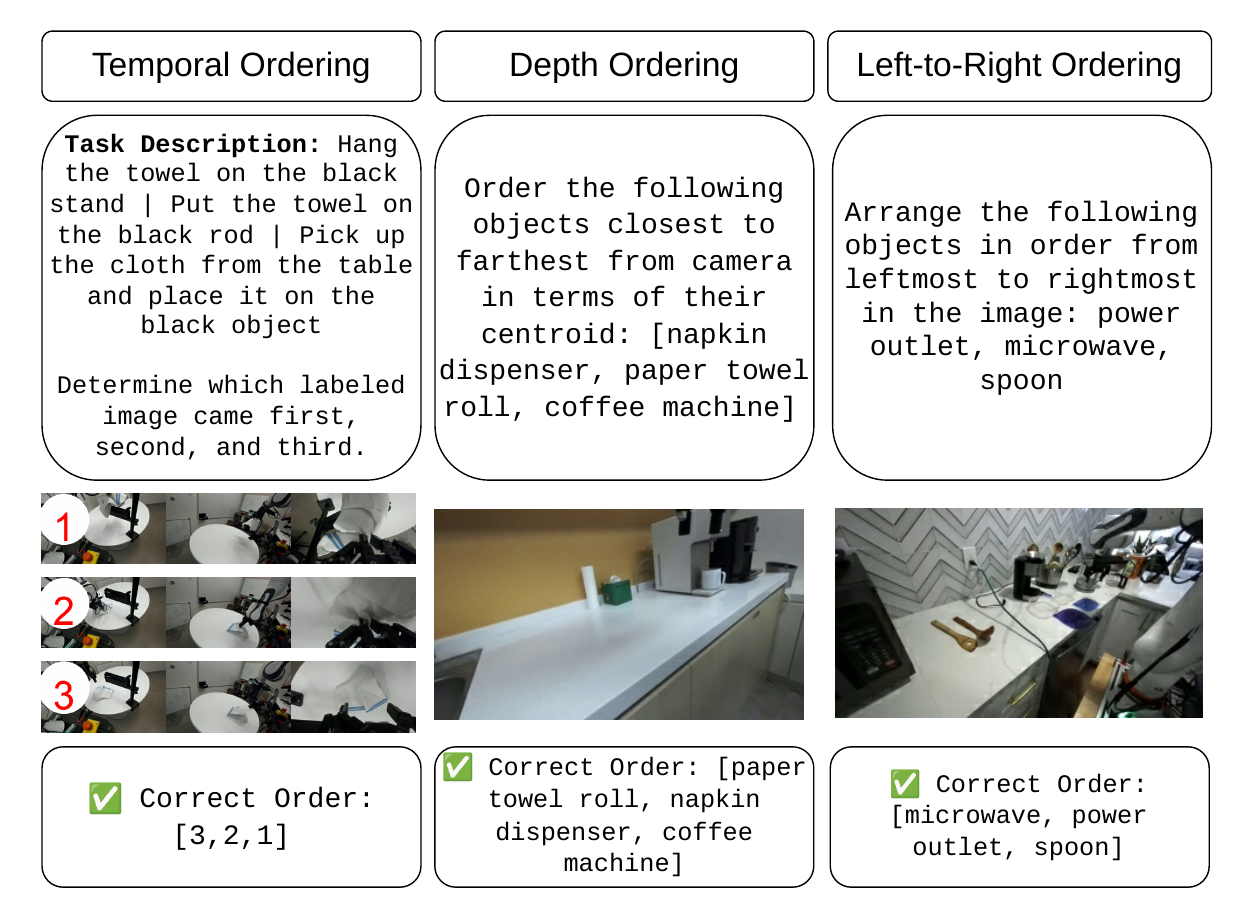}
\caption{Examples of spatiotemporal ordering tasks in our dataset.}
\label{fig:example_tasks}
\end{figure*}

Vision-Language Models (VLMs) have been adapted for diverse roles in robot manipulation pipelines as their capabilities have increased, serving as task planner components~\cite{yang2025guiding}, reward models~\cite{rocamonde2023vision}, action annotators~\cite{hancock2025actions}, and code policy generators~\cite{liang2022code}. 
Effectively performing these tasks requires sophisticated reasoning over large collections of images. Consider using a VLM to predict task progress based on recent observations; the VLM must be able to reason about spatial relationships between objects both within a frame and across frames---e.g., to measure whether the gripper moved closer to a target object to be grasped.\blfootnote{Correspondence to co-first author: \texttt{sanupam<at>engineering.upenn.edu}. Equal authorship is indicated with \textsuperscript{*} while \textsuperscript{\textdagger} indicates equal advising. Code and data available at \url{https://sagnikanupam.
com/papers/roboreason}.}

As a consequence, there is an urgent need to study and improve the fundamental capabilities of VLMs at spatiotemporal reasoning. While there has been significant recent interest in this problem~\cite{wang2025spatial457}, much of it focuses on general settings that do not capture the specific challenges that arise in robotics settings: (1) cluttered scenes from robot-centric viewpoints, (2) reasoning about object relationships across time, and (3) integrating information across multiple viewpoints.

We first conduct a systematic evaluation of multiple large VLMs on realistic spatiotemporal ordering tasks under a variety of settings, like having access to computer vision model outputs or having structured test-time compute. Our ordering problems are constructed from robot demonstration videos in the DROID dataset \cite{khazatsky2024droid} (examples in Figure \ref{fig:example_tasks}). Our results indicate that leading VLMs underperform on our dataset of real-world robotics images when compared to previously reported results. To address this performance gap, we present RoboReason, a scalable data generation pipeline for generating spatiotemporal reasoning training data from robot manipulation trajectories. We train much smaller models on a small amount (700 samples) of pipeline-generated data using Supervised Finetuning (SFT) and Group Relative Policy Optimization (GRPO) \cite{shao2024deepseekmath} and show that model performance increases significantly, even matching the performance of much larger models on some tasks. We also train small models on larger pipeline-generated training datasets of 8.8K samples and find that model performance also improves on downstream spatial and robotics reasoning tasks, demonstrating the potential for our approach to improve the efficacy of VLMs targeting robotics tasks.

\noindent Our main findings are as follows:
\begin{itemize}[leftmargin=*]
\item \textbf{VLMs that perform well on existing spatial reasoning benchmarks struggle on multi-object ordering tasks in real-world robot manipulation images.} We evaluate state-of-the-art embodied reasoning VLMs like Gemini Robotics-ER 1.5, GPT-5, and Qwen3-VL-235B on our tasks and discover that their performance decreases sharply as we increase the number of items to order. Additionally, GPT-5 and Gemini-ER-1.5 are unable to achieve more than 80\% accuracy on the easiest base setting of ordering two objects in any of our spatiotemporal tasks, despite reporting higher results in several other spatial reasoning benchmarks \cite{ray2024sat, tong2024cambrian, gemini2025geminirobotics15}.

\item \textbf{VLMs improve their performance in spatial ordering tasks with outputs from computer vision models.} We show that using a specialized vision model to predict segmentation maps or bounding boxes and then providing this information to the VLM mostly improves its accuracy on our left-to-right and depth-wise spatial ordering tasks. These results suggest that VLMs struggle with basic capabilities such as identifying objects in cluttered environments, compared to specialized vision models. %

\item \textbf{VLMs do not improve performance on temporal ordering tasks when provided with structured test-time compute, but do improve performance on spatial ordering tasks when provided with the same compute.} %
We study if explicitly structuring test-time compute by applying merge sort over VLM-generated pairwise comparisons aids performance in high reasoning settings. Our results show that merge sort performs worse or at a comparable level to providing all set elements to the VLM at once on temporal ordering tasks, but does improve spatial ordering performance. These results suggest that VLMs successfully reason over more than two frames to solve our temporal ordering tasks. %

\item \textbf{Both reinforcement learning and supervised finetuning (SFT) on pipeline-generated data improve small model performance on downstream reasoning tasks.} We scale our data construction pipeline to construct 8857 training samples for each ordering task, and finetune Qwen3VL-8B-Instruct on generated data. We observe that training on generated data improves performance on multiple downstream tasks. However, no single approach improves performance across all downstream tasks, and, in some cases, performance degrades after training on generated data.

\item \textbf{Reinforcement learning yields limited benefits on temporal ordering but offers large boosts on spatial ordering on training with a small set of generated data.} We perform reinforcement-learning finetuning experiments on Qwen2-VL-2B-Instruct and SmolVLM2-2.2B-Instruct using GRPO to maximize ordering performance on our tasks can improve its performance. Performance improves significantly on both 2D and depth spatial ordering tasks (with Qwen2-VL outperforming all larger VLMs on some tasks), while we observe more limited performance improvements on our temporal ordering task. Small model performance also increases proportional to the size of the RoboReason-generated training dataset. %

\end{itemize}

\noindent \textbf{Related work.} While several large-scale robot manipulation datasets have been released with associated robot camera images \cite{10611477, khazatsky2024droid}, attempts to utilize such trajectories for evaluating VLM reasoning have focused on trajectory-specific spatial orientation and robot-object interactions \cite{chen2025robo2vlm}. Alternatively, most approaches to evaluation VLM spatiotemporal reasoning use question-answer benchmarks with high-resolution images and answer options \cite{hudson2019gqa, antol2015vqa, wang2025spatial457}, which means VLM reasoning results can be influenced by guesswork, selection bias, and lack of visual noise or cluttered scenes \cite{balepur2024artifacts, zheng2023large, song2025robospatial}. Our approach instead leverages tasks using real-world robot camera images and trains models on multi-object spatiotemporal ordering tasks, decreasing the evaluation bias. We discuss related work in detail in Appendix \ref{app:rel_works}.

\section{RoboReason}

Our pipeline comprises of methods to construct three ordering tasks. In each task, the VLM is provided with a shuffled set of items alongside any task-specific context, and is tasked with reproducing the original ordering of the items. We use the pipeline task outputs as-is for our training experiments, but our test set has been checked by humans. We formalize the three ordering tasks as follows:

\begin{enumerate}
\item \textbf{Temporal ordering:} We assume we are provided with a set of video frames $f_1, f_2, \dots, f_n$ with associated timestamps $t_1, t_2, \dots, t_n$ and a language description $l$ describing the task being performed. After providing only the frames and the language description to the VLM, we ask the VLM to output a permutation of the frames $VLM(f_1, \dots, f_n; l) = i_1, \dots, i_n$ such that $t_{i_1} < t_{i_2} < \dots < t_{i_n}$. In the canonical task setting, each video frame comprises three images (left, right, and wrist camera) stitched into a single image.
\item \textbf{Depth ordering:} This task provides a single video frame $f$ alongside a list of the names of objects present in $f$ specified via their names. We denote the list of objects as $o_1, o_2, \dots, o_n$, and their associated distances from the camera $d_1, \dots, d_n$. The VLM is then prompted to output a permutation of the objects $VLM(o_1, \dots, o_n; f) = i_1, \dots, i_n$ such that $d_{i_1} < d_{i_2} < \dots < d_{i_n}$. 
\item \textbf{Left-to-right ordering:} This task is constructed in a similar way as depth ordering, except $d_1, \dots, d_n$ now measure the distance between the left edge of frame $f$ and the centroid of the object in 2D image space.
\end{enumerate}

\subsection{Data Generation Pipeline} 

\textbf{Temporal Ordering}. We randomly sample  trajectories from the DROID dataset \cite{khazatsky2024droid} that have language annotations. We divide each selected trajectory into 5 equal segments, and select one frame at random from each segment. We label the five images 1-5 after sorting in increasing temporal order, and for the corresponding 3 image setting, we select images 1, 3 and 5. %
Ground truth orderings are calculated using the timestamps of the trajectory provided in the original DROID trajectory. Recall that these are expert demonstration trajectories, so frames sampled from later in a trajectory will typically exhibit higher task progression than those from earlier in the trajectory. It is the ability to assess this robotics task progression that our ordering task is designed to evaluate.

\textbf{Depth Ordering}: Using the trajectories sampled for the temporal ordering task, we use the left camera view of a randomly selected video frame from each trajectory for the depth ordering task, since the wrist camera view is often directly above an object and does not show multiple objects at different depths. We identify objects present in the image with GPT-5, and construct bounding boxes using GroundingDINO \cite{liu2024grounding} for all identified objects. Then, we filter out all trajectories where fewer than 5 objects  with non-overlapping intersection over union mask, bounding boxes, and containment are visible in the scene. We impose a mask intersection over union (IoU) threshold of 0.01, a bounding box IoU threshold of 0.3, and a bounding box containment threshold of 0.5. If more than 5 such objects are present in a trajectory, we uniformly sample objects to initialize our set of 5 objects. For ground truth depth ordering, we generate depth maps for each image using Marigold \cite{ke2025marigold}, a monocular depth estimation model. We then generate segmentation masks of each object using Grounded SAM 2 \cite{ren2024grounded} (using the GPT-5 generated object names), and compute the mean depth of each object over all pixels in the segmentation mask.

\textbf{Left-to-Right Ordering}: For our left-to-right ordering task, we use the left-view camera images from the depth task, each from a unique DROID trajectory. For each image, similar to depth ordering above, we use the list of objects identified by GPT-5, and infer object bounding boxes for each with GroundingDINO. We first remove any objects we classify as scene objects (objects with $\texttt{width}_\texttt{bbox} > 0.7 \times \texttt{width}_\texttt{image}$). Then, we find the largest cluster of objects lying along a similar horizontal level, enforcing a maximum centroid y-coordinate difference of 0.5. Finally, in this cluster, we remove any objects that almost fully overlap with another object, i.e., have bounding boxes with $\text{IOU} > 0.7$. If we have less than 5 valid objects at this point, we discard the trajectory. The ground truth left-to-right ordering is constructed for valid trajectories using the x-coordinate of the centroid of the bounding box encompassing each object.

For the left-to-right and depth datasets, once we have filtered trajectories from the original temporal dataset, we continue sampling trajectories until we achieve the desired training set size. %

\subsection{Reinforcement Learning Finetuning}

To examine if reinforcement learning can be used as a viable method for improving model performance on our spatiotemporal ordering tasks, we train Qwen2-VL-2B-Instruct and SmolVLM2-2.2B-Instruct using GRPO \citep{shao2024deepseekmath} on our three ordering tasks using data generated via our pipeline. We use the Kendall rank correlation coefficient, defined as $\tau = 1-{\frac {2({\text{number of discordant pairs}})}{{n \choose 2}}}$, between the LLM generated rank and the ground truth rank as a reward function for all our tasks. %

\begin{table*}[t]
\tiny
\centering
\begin{tabular}{c|c|ccc|ccc|ccc}
\toprule
\multirow{2}{*}{Task Format} & \multirow{2}{*}{Model Name} & \multicolumn{3}{c}{Base Image} & \multicolumn{3}{c}{Segmentation Map Image} & \multicolumn{3}{c}{Bounding Box Image}
\\
& & Low & Medium & High & Low & Medium & High & Low & Medium & High \\
\midrule
\multirow{3}{*}{5 Image Set} & GPT-5 & 15.67 & \textbf{19.67} & 18.67 & 11.00 & \textbf{12.00} & 11.33 & 8.33 & 8.67 & \textbf{10.00} \\ 
& GPT-5-Nano & 0.67 & 2.33 & 1.00 & 2.00 & 2.33 & 0.67 & 1.00 & 3.33 &     2.33 \\
& Gemini-ER-1.5 & 1.33 & 6.67 & 7.33 & 3.67 & 4.00 & 5.33 & 2.33 & 4.33 & 5.33 \\
& Qwen3-VL-235B & 6.67 & 9.67 & 7.67 & 3.00 & 4.00 & 4.33 & 2.67 & 2.67 & 3.33 \\
\midrule
\multirow{3}{*}{3 Image Set} & GPT-5 & \textbf{49.33} & 47.67 & 47.33 & 44.67 & \textbf{45.00} & 41.33 & 39.67 & 38.67 & \textbf{40.67} \\
& GPT-5-Nano & 17.00 & 23.33 & 23.67 & 20.33 & 20.00 & 24.67 & 18.33 & 22.67 & 22.00 \\
& Gemini-ER-1.5 & 38.33 & 36.67 & 38.00 & 32.67 & 37.33 & 35.67 & 32.00 & 35.00 & 34.67 \\
& Qwen3-VL-235B& 32.33 & 32.00 & 32.67 & 30.33 & 28.00 & 28.67 & 29.67 & 28.00 & 27.00 \\
\bottomrule
\end{tabular}
\caption{Model Accuracies (\%) for Temporal Ordering Tasks.}
\label{tab:temporal_ordering}
\end{table*}

\begin{table*}[t]
\centering
{\tiny
\begin{tabular}{c|c|ccc|ccc|ccc}
\toprule
\multirow{2}{*}{Task Format} & \multirow{2}{*}{Model Name} & \multicolumn{3}{c}{Base Image} & \multicolumn{3}{c}{Segmentation Map Image} & \multicolumn{3}{c}{Bounding Box Image} \\
& & Low & Medium & High & Low & Medium & High & Low & Medium & High \\
\midrule
\multirow{3}{*}{5 Object Set} & GPT-5 & 4.33 & 4.33 & 5.00 & \textbf{9.33} & 8.00 & 6.67 & 7.67 & \textbf{10.33} & 7.33 \\
& GPT-5-Nano & 2.33 & 4.33 & 3.00 & 5.67 & 3.33 & 2.67 & 3.00 & 3.33 & 3.67 \\
& Gemini-ER-1.5 & 3.00 & \textbf{6.00} & 5.67 & 4.33 & 8.00 & 7.33 & 6.33 & 9.33 & 9.33 \\
& Qwen3-VL-235B & 2.67 & 3.00 & 3.33 & 5.00 & 3.00 & 4.00 & 5.33 & 5.00 & 4.67 \\
\midrule
\multirow{3}{*}{3 Object Set} & GPT-5 & 34.00 & 34.00 & 32.67 & \textbf{42.33} & \textbf{42.33} & 41.33 & 43.67 & 41.67 & 39.00\\
& GPT-5-Nano & 26.00 & 25.67 & 25.67 & 30.67 & 29.00 & 26.67 & 27.00 & 29.67 & 26.67 \\
& Gemini-ER-1.5 & 25.67 & \textbf{35.67} & \textbf{35.67} & 31.00 & 37.33 & 38.33 & 38.33 & 40.00 & \textbf{44.00} \\
& Qwen3-VL-235B & 29.67 & 30.00 & 29.67 & 31.67 & 30.67 & 30.67 & 34.67 & 34.00 & 36.00 \\
\bottomrule
\end{tabular}
}
\caption{Model Accuracies (\%) for Depth Ordering Tasks}
\label{tab:depth_ordering}
\end{table*}

\begin{table*}[t]
\tiny
\centering
\begin{tabular}{c|c|ccc|ccc|ccc}
\toprule
\multirow{2}{*}{Task Format} & \multirow{2}{*}{Model Name} & \multicolumn{3}{c}{Base Image} & \multicolumn{3}{c}{Segmentation Map Image} & \multicolumn{3}{c}{Bounding Box Image}
\\
& & Low & Medium & High & Low & Medium & High & Low & Medium & High \\
\midrule
\multirow{3}{*}{5 Object Set} & GPT-5 & 21.00& 21.00& 21.00& 17.67 & 19.33 & 19.67 & 20.33 & 18.00& 22.67 \\
& GPT-5-Nano & 7.00& 8.67 & 11.33 & 5.33 & 7.67 & 10.33 & 4.41 & 5.67 & 7.33 \\
& Gemini-ER-1.5 & 18.79 & 18.06 & 19.38 & 17.58 & 15.71 & 18.37 & 28.77 & \textbf{30.64} & 27.21 \\
& Qwen3-VL-235B & \textbf{22.33} & 22.00& 19.00& 14.33 & 17.67 & \textbf{23.53} & 20.67 & 18.33 & 21.00\\
\midrule
\multirow{3}{*}{3 Object Set} & GPT-5 & 58.00& 57.00& \textbf{60.00} & 56.33 & 55.00& \textbf{57.00} & 49.67 & 52.00& 52.67 \\
& GPT-5-Nano & 39.67 & 43.00& 44.67 & 36.67 & 40.67 & 42.00& 33.33 & 36.00& 41.00\\
& Gemini-ER-1.5 & 56.00& 56.23 & 59.12 & 50.84 & 55.82 & 56.8 & 61.67 & \textbf{64.77} & 62.96 \\
& Qwen3-VL-235B & 57.00& 58.33 & 58.67 & 50.00& 50.00& 48.33 & 56.00& 55.33 & 61.8 \\
\bottomrule
\end{tabular}
\caption{Model Accuracies (\%) for Left-to-Right Ordering Tasks}
\label{tab:left_to_right_ordering}
\end{table*}

\begin{table*}[t]
\tiny
\centering
\tiny
\begin{tabular}{c|c|ccc|ccc|ccc}
\toprule
\multirow{2}{*}{Task Format} & \multirow{2}{*}{Model Name} & \multicolumn{3}{c}{Temporal Ordering Task} & \multicolumn{3}{c}{Depth Ordering Task} & \multicolumn{3}{c}{Left-to-Right Ordering Task}
\\
& & Low & Medium & High & Low & Medium & High & Low & Medium & High \\
\midrule
\multirow{3}{*}{5 Image/Object Set} & GPT-5 & 15.67 & \textbf{19.67} & 18.67 & 4.33 & 4.33 & 5.00 & 21.00 & 21.00 & 21.00 \\
& GPT-5-Nano & 0.67 & 2.33 & 1.00 & 2.33 & 4.33 & 3.00 & 7.00 & 8.67 & 11.33 \\
& Gemini-ER-1.5 & 1.33 & 6.67 & 7.33 & 3.00 & \textbf{6.00} & 5.67 & 18.79 & 18.06 & 19.38 \\
& Qwen3-VL-235B & 6.67 & 9.67 & 7.67 & 2.67 & 3.00 & 3.33 & \textbf{22.33} & 22.00 & 19.00 \\
\midrule
\multirow{3}{*}{4 Image/Object Set} & GPT-5 & 27.00 & 29.33 & \textbf{30.00} & 9.67 & 13.00 & 11.00 & 32.67 & 32.00& 37.00\\
& GPT-5-Nano & 6.67 & 8.00 & 5.33 & 6.67 & 9.33 & 8.33 & 17.67 & 23.67 & 23.67 \\
& Gemini-ER-1.5 & 16.33 & 23.67 & 18.67 & 11.00 & 14.67 & 11.00 & \textbf{38.41} & 34.93 & 37.25 \\
& Qwen3-VL-235B & 18.33 & 16.00 & 13.67 & 17.67 & \textbf{18.33} & 17.00 & 37.67 & 38.33 & 34.33 \\
\midrule
\multirow{3}{*}{3 Image/Object Set} & GPT-5 & \textbf{49.33} & 47.67 & 47.33 & 34.00 & 34.00 & 32.67 & 58.00& 57.00& \textbf{60.00}  \\
& GPT-5-Nano & 17.00 & 23.33 & 23.67 & 26.00 & 25.67 & 25.67 & 39.67 & 43.00& 44.67 \\
& Gemini-ER-1.5 & 38.33 & 36.67 & 38.00 & 25.67 & \textbf{35.67} & \textbf{35.67} & 56.00& 56.23 & 59.12 \\
& Qwen3-VL-235B & 32.33 & 32.00 & 32.67 & 29.67 & 30.00 & 29.67 & 57.00& 58.33 & 58.67\\
\midrule
\multirow{3}{*}{2 Image/Object Set} & GPT-5 & 64.00 & \textbf{67.67} & 67.33 & 52.33 & 53.67 & 53.33 & 78.33 & \textbf{80.00} & 77.67 \\
& GPT-5-Nano & 53.33 & 52.00 & 52.00 & 52.00 & 49.67 & 49.33 & 63.33 & 67.00& 69.00\\
& Gemini-ER-1.5 & 50.67 & 58.33 & 60.33 & 49.33 & 54.67 & 54.33 & 74.67 & 78.26 & 75.59 \\
& Qwen3-VL-235B & 60.67 & 62.33 & 59.67 & 70.00 & 68.00 & \textbf{71.33} & 78.67 & 78.00& 78.67 \\
\bottomrule
\end{tabular}
\caption{Model Accuracies (\%) Across Different Set Cardinalities. All results are in the base setting (i.e. no images had any bounding boxes or segmentation map overlays on them).}
\label{tab:set_lengths}
\end{table*}

\section{Experiments}
\label{sec:experiments}
We conduct two sets of experiments. First, we evaluate the spatiotemporal reasoning ability of current state-of-the-art VLMs (Section \ref{sec:zero_shot}). Second, we evaluate our training data generation pipeline by human validation of the pipeline outputs and evaluating small VLMs finetuned on the pipeline-generated data (Section \ref{sec:roboreason_evals}).

Our experiments seek to answer the following questions with respect to large VLM performance:
\begin{enumerate}
\item How does VLM spatiotemporal ability vary across different reasoning levels? %
\item Does structuring test-time compute by performing pairwise comparisons and using merge sort help improve model performance? %
\item At what rate does model performance decline as more objects/images are added to the ordering list, and how does this rate vary for different models? %
\item Does overlaying segmentation maps and bounding boxes on images help improve VLM spatiotemporal reasoning performance? %
\end{enumerate}

We attempt to answer the following questions to evaluate RoboReason-generated training data:
\begin{enumerate}
\item How accurate are the pipeline generated outputs with respect to human annotations? %
\item Does finetuning on our ordering tasks improve downstream robotics or visual reasoning task performance? %
\item How does finetuned model performance on spatiotemporal ordering change with i) training task type and ii) amount of pipeline-generated data used?
\end{enumerate}

\subsection{Experimental Setup} 

\begin{table*}[htbp]
\tiny
\centering

\end{table*}

Since we seek to evaluate models under different reasoning levels, we set output token limits for low, medium, and high reasoning settings in our tasks (described in  Table \ref{tab:reasoning_levels}), %
based on the maximum output tokens generated by GPT-5 in a given OpenAI reasoning setting (low, medium, high) for a particular ordering task. This helps ensure a consistent comparison with other models, which require specifying reasoning or output budgets in tokens. For the merge sort format of the tasks (which we only evaluate in the high reasoning setting), the token counts are determined separately using the maximum token count produced by GPT-5 while performing pairwise comparisons.

For our three ordering tasks, we present results from 3 settings: the base image  setting (which only provides the VLM with the original DROID image), the segmentation map setting (which overlays the GroundedSAM 2 segmentation map associated with each object on top of the image), and the bounding box setting (which overlays the GroundingDINO bounding box associated with each object on top of the image). For depth and left-to-right ordering tasks, we only overlay bounding boxes and segmentation maps associated with the objects to be ordered. For the temporal ordering setting, we overlay bounding boxes and segmentation maps on all 3 camera views for objects that GPT-5 identifies as relevant objects from the task description. We present our results for temporal ordering in Table \ref{tab:temporal_ordering}, depth ordering in Table \ref{tab:depth_ordering}, and left-to-right ordering in Table \ref{tab:left_to_right_ordering}. In the base image setting, we examine how the number of objects to order affects performance in Table \ref{tab:set_lengths}, where we randomly sample images/objects from our initial 5-length image/object set to construct the 2 and 4 length versions for all task types. In all our ordering experiments, we calculate accuracy as the percentage of samples where the model outputs exactly the same order as the ground truth values. 
For our small generated dataset experiments, we construct a dataset of 1000 samples for each task type. A subset of 300 randomly sampled tasks are then selected as a test set, with the remaining being used as our small training dataset. For the depth and left-to-right test sets, since the ground truths are model generated, we validate the pipeline's generated object labels and ordering by having a human correct any ambiguous labels or incorrect orderings. All our ordering task results are from the human-validated, 300 sample test set. For our downstream task validation experiments, we scale the training dataset to over 8.8K samples for each task type. 

We evaluate our finetuned checkpoints on visual reasoning tasks using the SAT \citep{ray2024sat} benchmark, as well as two downstream robotics tasks: autoregressive value estimation using the RT-1 \cite{brohan2022rt} and Dobb-E \cite{shafiullah2023bringing} datasets (proposed in  \citep{ma2024vision}), and robotics failure analysis question-answering using the real-world trajectory question-pairs from the RoboFAC dataset \citep{ye2025robofac}. 

\subsection{Zero-Shot VLM Performance}
\label{sec:zero_shot}
\begin{table}[H]
\tiny
\centering
\begin{tabular}{c|ccc}%
\toprule 
Model Name & \multicolumn{3}{c}{Across All 3 and 5 Set Cardinality Ordering Tasks}  \\
& Base & Segmentation Map & Bounding Box \\ %
\midrule
GPT-5 & 1.83 $\pm$ 1.49 & 2.06 $\pm$ 1.02 & \textbf{3.17} $\pm$ 1.28\\
GPT-5-Nano & 3.33 $\pm$ 2.39 & \textbf{3.94} $\pm$ 1.39 & 3.49 $\pm$ 2.37\\
Gemini-ER-1.5 & 3.96 $\pm$ 3.43 & \textbf{3.99} $\pm$ 2.37 & 3.53 $\pm$ 1.06\\
Qwen3-VL-235B & 1.61 $\pm$ 1.29 & \textbf{2.92} $\pm$ 3.11 & 2.52 $\pm$ 2.14\\
\bottomrule
\end{tabular}
\caption{Mean and Standard Deviation of Reasoning Difference}
\label{tab:reasoning_diff_mean}
\end{table}

\begin{table}[H]
\tiny
\centering
\begin{tabular}{c|c|ccc}
\toprule
Task Format & Model Name & Temporal & Depth & Left-to-Right
\\
& & Ordering & Ordering & Ordering \\
\midrule
\multirow{3}{*}{5 Image/Object Set} & GPT-5 & \textbf{8.00} & 5.67 & \textbf{23.00} \\
& GPT-5-Nano & 1.00 & 5.00 & 12.67 \\
& Gemini-ER-1.5 & 1.67 & \textbf{6.33} & 22.33 \\
& Qwen3-VL-235B & 0.33 & 3.33 & 22.00 \\
\midrule
\multirow{3}{*}{3 Image/Object Set} & GPT-5 & \textbf{42.33} & \textbf{33.00} & 60.00 \\
& GPT-5-Nano & 23.00 & 27.00 & 47.33 \\
& Gemini-ER-1.5 & 16.33 & 31.33 & 59.00 \\
& Qwen3-VL-235B & 20.33 & 23.00 & \textbf{61.00} \\
\bottomrule
\end{tabular}
\caption{Model Accuracies (\%) for High-Reasoning Merge Sort Based Ordering in Base Image Setting}
\label{tab:merge_sort}
\end{table}

\textbf{Impact of Reasoning Levels:}
To measure the impact of reasoning levels, we measure the difference in task performance in a particular setting (for example, 3 image base setting temporal ordering) between the best and worst performing reasoning level. We refer to this quantity as the reasoning difference, and report its mean and standard deviations across different tasks in Table \ref{tab:reasoning_diff_mean}.

While in most settings, higher reasoning levels are better, in the base setting, we sometimes find that lower reasoning levels outperform higher ones, especially in 2 object/image temporal and depth settings. We hypothesize that this is because the low reasoning level does not allow for enough tokens to reason about ordering a large number of images or objects, while for pairwise comparisons, generating excessive tokens can lead the model to ``overthink'' and draw inaccurate conclusions.

\textbf{Impact of Structured Test-Time Compute:} 
Given the relatively high performance of VLMs while ordering pairs of objects, we evaluate the performance of VLMs by scaling their test time compute using an external structure. We implement a merge sort algorithm that calls the VLM to make pairwise comparisons, and our results (included in Table \ref{tab:merge_sort}) show that in the high reasoning setting on the spatial ordering tasks, VLMs usually perform at least as good as and often better in the structured test-time compute approach than when provided with all objects to order at once. On the other hand, they perform much better on temporal tasks when they are provided with all the frames at once, suggesting that VLMs improve their performance by reasoning over multiple images correctly in the temporal ordering task, and some of this information is lost with pairwise comparisons.

\begin{figure*}[t]
\centering
\resizebox{\textwidth}{!}{%
\begin{tabular}{cc}
\begin{subfigure}[t]{0.45\linewidth}
\centering
\includegraphics[width=\linewidth]{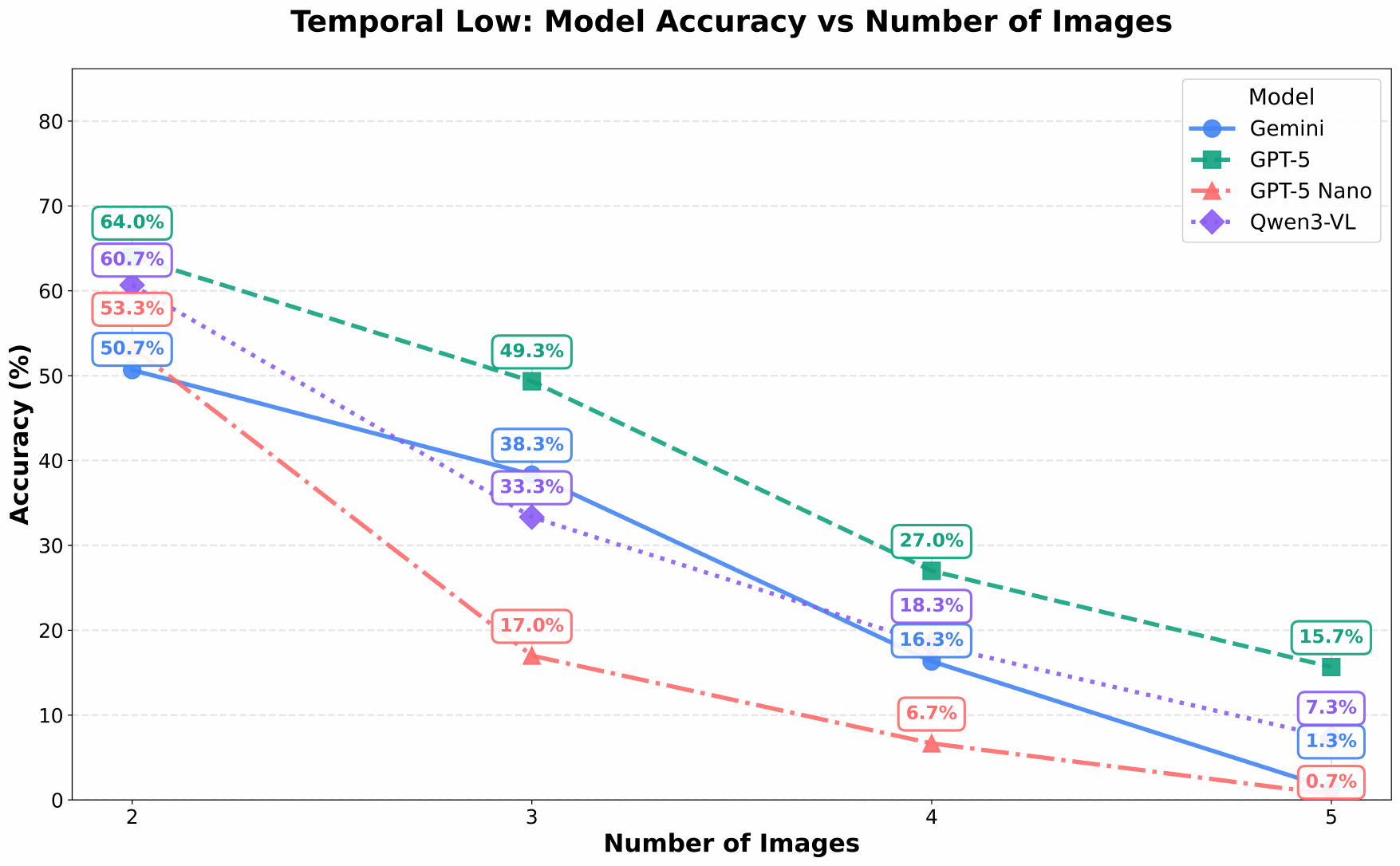}
\caption{Temporal Low}
\label{fig:temporal_low}
\end{subfigure}
&
\begin{subfigure}[t]{0.45\linewidth}
\centering
\includegraphics[width=\linewidth]{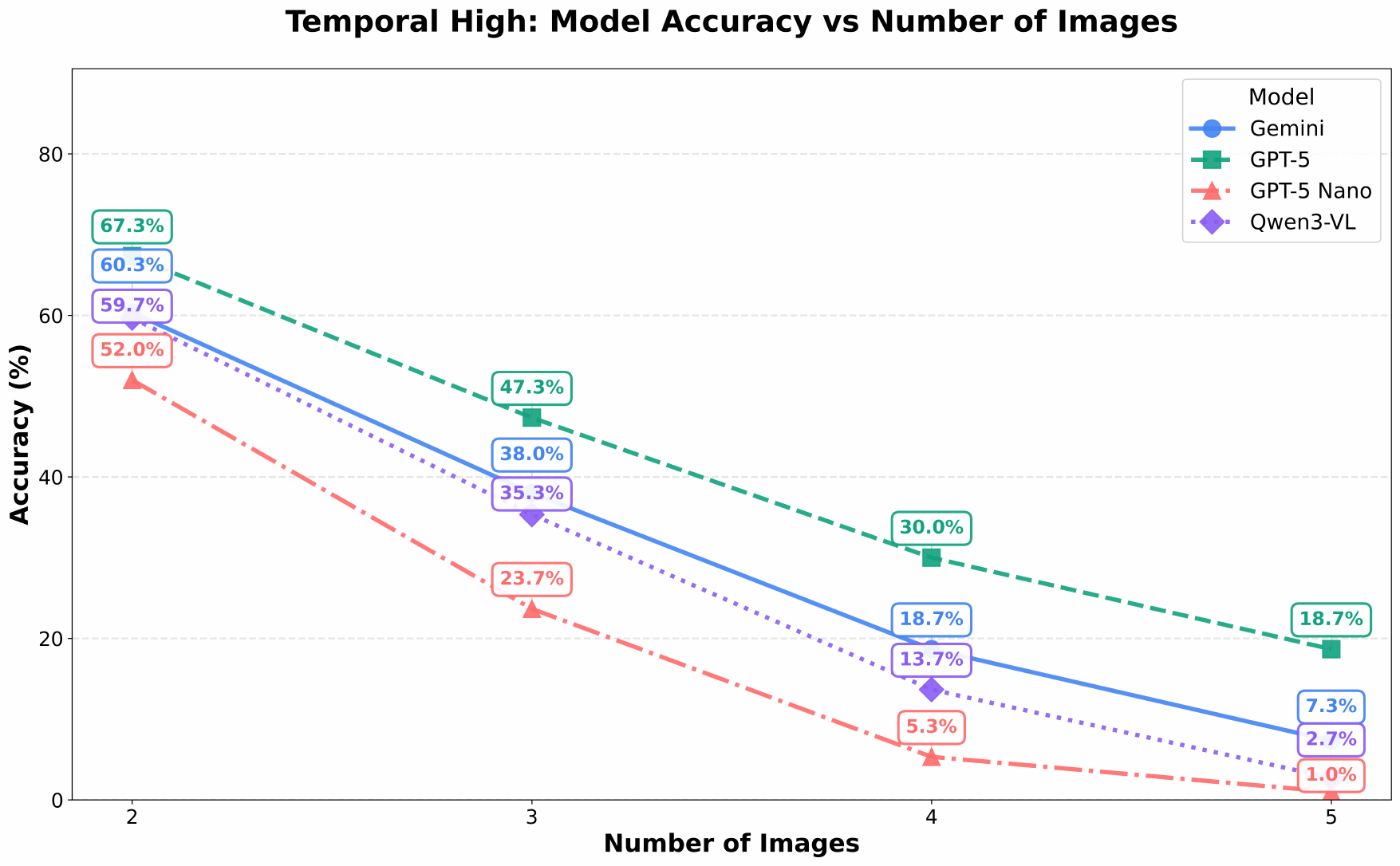}
\caption{Temporal High}
\label{fig:temporal_high}
\end{subfigure}
\\[0.5em]
\begin{subfigure}[t]{0.45\linewidth}
\centering
\includegraphics[width=\linewidth]{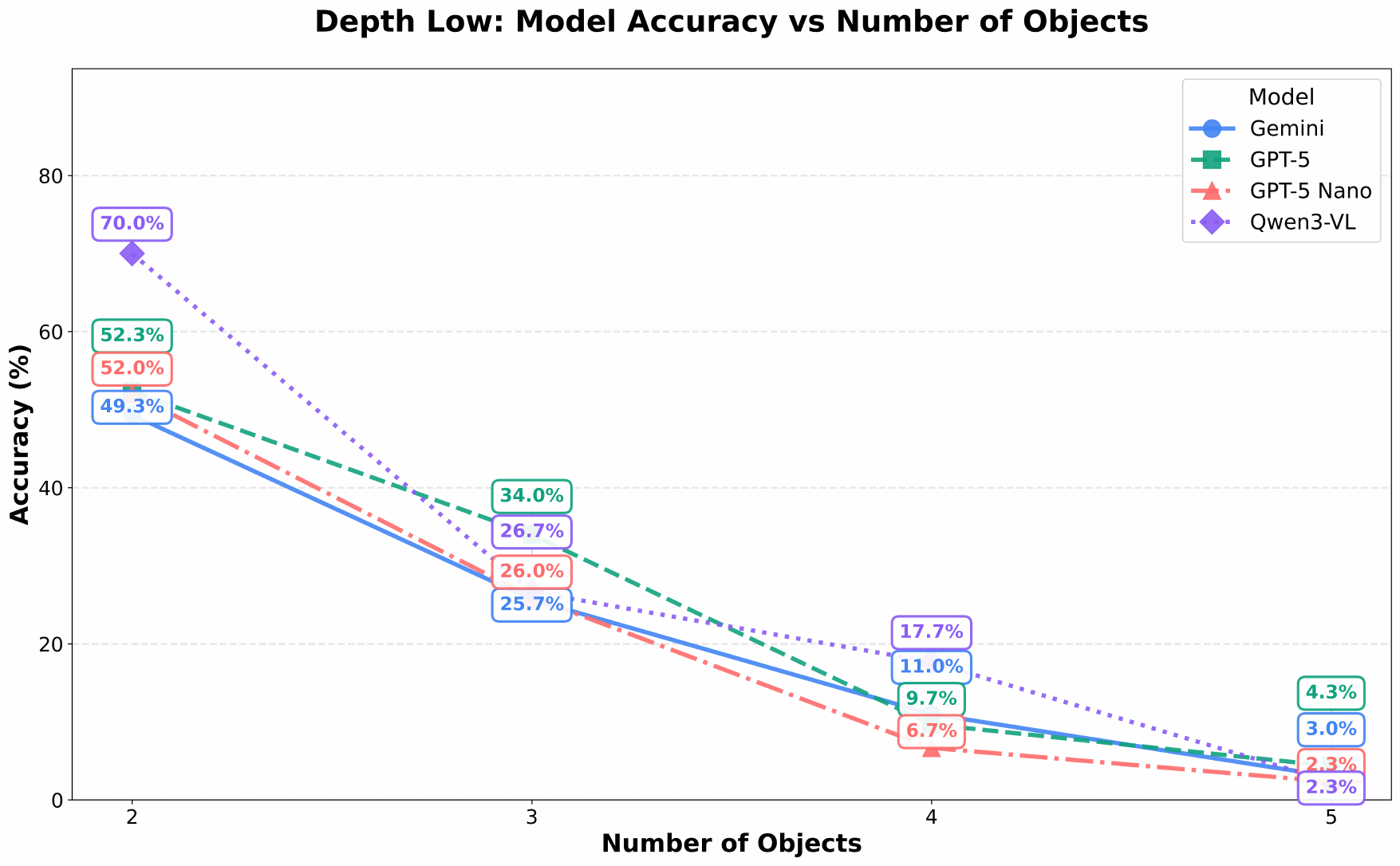}
\caption{Depth Low}
\label{fig:depth_low}
\end{subfigure}
&
\begin{subfigure}[t]{0.45\linewidth}
\centering
\includegraphics[width=\linewidth]{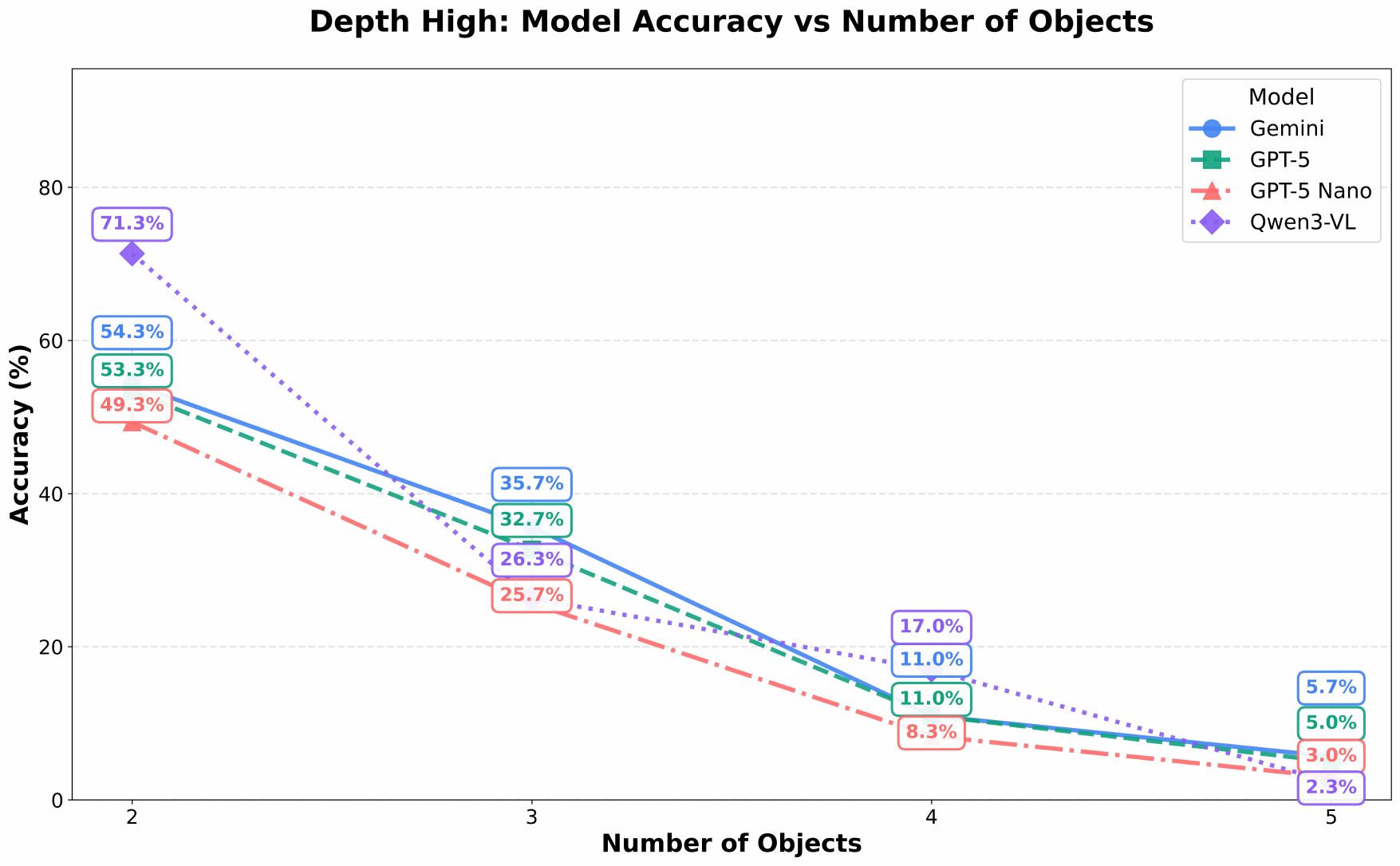}
\caption{Depth High}
\label{fig:depth_high}
\end{subfigure}
\end{tabular}
}
\caption{Comparison of how model performance scales with set cardinality in temporal and depth ordering tasks in the base setting. In the temporal high reasoning setting, model performance scales predictably with increase in set cardinality, maintaining a consistent relative order across all set cardinalities. However, the relative order is inconsistent in the remaining settings in the figure.}
\label{fig:temporal_depth_2345}
\end{figure*}

\textbf{Impact of Set Cardinalities:}
Model performance on spatiotemporal ordering tasks declines sharply as more elements are added to the set to be ordered. However, different VLMs decline at different rates, and in on certain task settings, two models (for example, GPT-5 and GPT-5-Nano with low reasoning in the depth ordering task) that perform at relatively similar levels when asked to order 2 elements have a much larger gap in performance when asked to order 3 or more elements. It also may be the case that models that perform at a similar or worse level as other models at smaller set cardinalities perform relatively better at larger set cardinalities, as evidenced by GPT-5 in the aforementioned setting when the models are asked to order 3 or more objects. We show some examples of these occurrences in Figure \ref{fig:temporal_depth_2345}.

\textbf{Impact of Segmentation Map and Bounding Box Overlays:}
For our two canonical settings (ordering 5 and 3 images/objects), we study the difference in performance caused by passing images overlaid with segmentation maps and bounding boxes. Bounding boxes achieve comparable (within 2\% of original accuracy) or better performance for all models at each reasoning level in the depth ordering task, and improve performance for Gemini-ER-1.5 and Qwen3-VL-235B in some reasoning levels in the left-to-right ordering tasks. Similarly, segmentation maps also achieve comparable or better performance for all models at each reasoning level in the depth ordering task, although they do not perform as well as bounding box overlays in the left-to-right ordering task.

Mean reasoning difference increases for all models on using images with overlaid segmentation maps, and increases for all models except Gemini-ER-1.5 on using overlaid bounding boxes. Additionally, using bounding box and segmentation map overlays also decreases the standard deviation slightly for all models except Qwen3-VL-235B. Our results suggest that using additional tokens when images are overlaid with computer vision model outputs may improve performance more than expending more tokens in the base setting.

Our results demonstrate that large VLMs have limited performance on spatiotemporal ordering tasks, which often falls short of even prior reported results when applied to our robotics-relevant data.

\subsection{RoboReason Evaluations}
\label{sec:roboreason_evals}

To address these limitations, we generate training data using the scalable RoboReason pipeline. We use it to finetune VLMs, focusing on smaller models due to compute limitations. For evaluating our training data pipeline, we first compare the pipeline outputs to those of a human evaluator. After finetuning small VLMs on the pipeline-generated data, we compare the post-trained models' performance to the original model on both our spatiotemporal ordering tasks and downstream robotics and visual reasoning tasks. Our results show that while the pipeline ground truth orders differ from human evaluator's assessments in a significant percentage of cases, our data generation pipeline still produces training data that provides a strong signal for learning and helps improve small model performance on both ordering and downstream tasks. 

\textbf{Human Evaluation:}
Since our depth and left-to-right ordering task ground truths are generated by computer vision models, we have a human evaluator validate the 5 object depth and left-to-right test sets prior to evaluating VLMs. The evaluator corrects the ordering if they identify all the objects in the image and determine the pipeline-generated ordering to be incorrect. If any labels are ambiguous, and the evaluator cannot identify an object in the image, they modify the object label prior to updating the ordering. Since the test set trajectories are a random subset of DROID trajectories (and thus drawn from the same distribution as the training set), we present our human evaluator disagreement results as an evaluation of our training data generation pipeline. %
For the left-to-right tasks, 15.67\% of samples in the test set were determined to have at least one object with an incorrect position in the ground truth order, and 29\% of samples were determined to have at least one ambiguous label. For the depth tasks, 26.67\% of samples in the test set were determined to have at least one object with an incorrect position in the ground truth order, while 22.67\% of samples were determined to have at least one ambiguous label.

\textbf{Impact of Training on Pipeline-Generated Data on Downstream Reasoning Tasks:} To investigate if training on RoboReason-generated data can improve performance on downstream reasoning tasks, we finetune a Qwen3VL-8B-Instruct model on 8.8K samples each for our three task types, using both SFT and GRPO. We evaluate each finetuned model on both our ordering tasks, as well as additional visual and robotics reasoning tasks, reporting our results in Table \ref{tab:downstream_tasks} and experimental setup details in Appendix \ref{app:large_scale_training_task}.

\begin{table*}[t]
\centering
\setlength{\tabcolsep}{4pt}
\resizebox{\textwidth}{!}{%
\begin{tabular}{c|ccc|ccccc}
\toprule
Model & \multicolumn{3}{c|}{3 Image/Object Spatiotemporal Ordering} & \multicolumn{5}{c}{Downstream Reasoning} \\
& Temporal & Depth & Left-to-Right & SAT & RoboFAC (MCQ) & RoboFAC (Open-Ended) & Value Est. (RT-1) & Value Est. (Dobb-E) \\
\midrule
Base                      & 18.00           & 22.33           & 53.67           & \textbf{70.67}         & 62.55           & 12.28           & 0.0066        & \textbf{0.0123}\\
SFT (Temporal)            & \textbf{50.67}  & 26.00           & 55.33           & \textbf{70.67}         & 60.21           & 12.92           & 0.0001   & 0.0114   \\
SFT (Depth)               & 22.67           & \textbf{31.67}  & 56.67           & 68.67 & 60.26           & \textbf{13.33}           & 0.0047 & 0.0015 \\
SFT (Left-to-Right) & 23.67 & 24.67 & \textbf{61.33} & 69.33 & 58.91 & 13.16 & 0.0080 & 0.0062\\
GRPO (Temporal)           & 22.67           & 26.00           & 51.67           & 68.67         & \textbf{63.13}  & 11.94           & 0.0060 & 0.0036\\
GRPO (Depth)%
& 22.67           & 31.33           & 51.00           & 68.67        & 59.84           & 8.28            & \textbf{0.0129}         & 0.0071         \\
GRPO (Left-to-Right)      & 20.67           & 17.33           & 55.33           & \textbf{70.67} & 56.25%
& 11.76%
& 0.0001 %
& 0.0000        \\
\bottomrule
\end{tabular}%
}
\caption{Finetuned Qwen3VL-8B Results on Spatiotemporal Ordering and Downstream Reasoning Tasks. Value estimation results are average value-order correlations (Appendix \ref{app:large_scale_training_task}), while all other results are question-answering accuracies (in \%).  %
}
\label{tab:downstream_tasks} %
\end{table*}

\textbf{Impact of Task Type and Amount of Data on Finetuning Performance:}
We show that finetuning on even a small amount of pipeline-generated data significantly improves performance on base image setting ordering tasks in Table \ref{tab:reinforcement_learning}, enabling the small models to reach the same level of performance as much larger models on the spatial ordering tasks while showing smaller gains on temporal ordering tasks (e.g. Qwen2-VL-2B surpasses the performance of all larger VLMs on the 3-object depth and left-to-right ordering tasks). In general, we find that reinforcement learning improves performance to a greater extent than supervised finetuning on the same task.
We also study the impact of the training data amount on the model performance, by training multiple models in each of our 3 Image/Object base image settings with different amounts of training data (Table \ref{tab:reinforcement_learning_300_700}). Our results show that model performance scales proportionally to training dataset size, suggesting that scaling training data using our pipeline can help models perform better at spatiotemporal ordering.

\begin{table}[H]
\tiny
\centering
\begin{tabular}{cccccc}
\toprule
Model & Task Format & Training & Temporal & Depth & Left-to- \\
Name & & Method & & & Right\\
\midrule
\multirow{6}{*}{Qwen2-VL-2B} & \multirow{3}{*}{5 Images/Objects} & Base & 1.00 & 0.66 & 0.67 \\ 
& & GRPO & \textbf{2.00} & \textbf{16.33} & \textbf{17.33} \\
& & SFT & 0.33 & 10.33 & 15.33 \\
& \multirow{3}{*}{3 Images/Objects} & Base & 17.33 & 15.66 & 16.00 \\ 
& & GRPO & \textbf{21.67} & \textbf{66.00} & \textbf{67.66} \\
& & SFT & 16.00 & 36.00 & 47.66 \\
\midrule
\multirow{6}{*}{SmolVLM2-2.2B} & \multirow{3}{*}{5 Images/Objects} 
& Base & 0.33 & 0.00 & 2.00 \\
& & GRPO & 1.33  & \textbf{2.00} & \textbf{16.66} \\
& & SFT & \textbf{1.66} & 1.00 & 9.33\\
& \multirow{3}{*}{3 Images/Objects} 
& Base & 13.00 & 5.33 & 12.33  \\
& & GRPO & \textbf{17.00} & \textbf{34.00} &  \textbf{58.67} \\
& & SFT & 12.67 & 19.33 & 44.00 \\
\bottomrule
\end{tabular}
\caption{Model Accuracies (\%) for GRPO and SFT}
\label{tab:reinforcement_learning}
\end{table}

\begin{table}[H]
\tiny
\centering
\begin{tabular}{ccccc}
\toprule
Model Name & Training Method & Temporal & Depth & Left-to-Right \\
\midrule
\multirow{3}{*}{Qwen2-VL-2B}
& Base & 17.33 & 15.66 & 16.00 \\ 
& GRPO-300 & 19.00 & 37.00 & 54.00 \\
& GRPO-500 & 20.00 & 56.00 & 63.66 \\
& GRPO-700 & \textbf{21.67} & \textbf{66.00} & \textbf{67.66} \\
\midrule
\multirow{3}{*}{SmolVLM2-2.2B} 
& Base & 13.00 & 8.00 & 12.33   \\
& GRPO-300 & 15.33 & 24.33 & 45.00 \\
& GRPO-500 & 15.66 & 30.66 & 53.33 \\
& GRPO-700 & \textbf{17.00} & \textbf{34.00} & \textbf{58.67} \\
\bottomrule
\end{tabular}
\caption{3 Image/Object ordering model accuracies (\%) after GRPO under different training data sizes}
\label{tab:reinforcement_learning_300_700}
\end{table}

\section{Conclusion}

We have introduced RoboReason, a scalable task generation pipeline for spatiotemporal reasoning that utilizes real-world robot manipulation images. Experimentally, we first establish the limits of today's VLMs: we demonstrate the impact of reasoning levels, ordering set cardinalities, structured test-time compue and computer vision tools on VLM spatiotemporal reasoning. %
Next, we demonstrate that by finetuning smaller VLMs on our pipeline-generated data, we can surpass the performance of much larger VLMs on spatial ordering tasks and improve model performance on downstream reasoning tasks. 
These results suggest that utilizing reinforcement learning approaches and computer vision model outputs are encouraging directions for future work on real-world spatiotemporal reasoning. We believe that RoboReason presents a promising way forward for improving robotics-relevant VLM capabilities.

\textbf{Limitations.}
For the depth and left-to-right ordering tasks, our pipeline can sometimes produce erroneous ground truths due to failures in individual modules such as object detection, bounding box generation, and depth estimation. Additionally, our filtering methods sometimes introduce ambiguity by selecting objects that may overlap or be equidistant from the camera. 
While using GPT-5 generated object labels may theoretically favor GPT-5 during evaluation due to egocentric bias, we do not see empirical evidence of this occurring, as in several instances, other models outperform it on spatial ordering tasks, and its own performance is sometimes hurt by providing it with images with overlays associated with the objects it had identified. For temporal ordering, it can sometimes be difficult to reason about task progression given certain task descriptions, since the VLM must estimate the direction of motion based only on static images of the scene. Such tasks constitute a minority of tasks in our temporal ordering dataset, and are often solvable using other environment context clues, but our pipeline does not explicitly filter for such ambiguous tasks descriptions. Finally, for temporal ordering tasks, we observe that VLMs struggle to reason across views (as shown in Appendix \ref{app:camera_view_impact}) and typically improve performance when restricted to a single camera view. However, since we wish to improve multi-view reasoning during training, we generate multi-view tasks in our temporal setting.

\section*{Acknowledgements}

This work is funded in part by the UPenn DARPA MOCHA CR PO-0074571/10101353 subcontract of Peraton Labs, DARPA TIAMAT HR0011249042, NSF CAREER 2239301, ONR N00014-22-1-2677, and NSF SLES 2331783. Yao Tang was supported in part by a gift from AWS to Penn Engineering’s ASSET Center for Trustworthy AI. Any opinions, findings and conclusions or recommendations expressed in this material are those of the author(s) and do not necessarily reflect the views of DARPA, NSF, Peraton Labs, or AWS.
\bibliography{custom}

@String(CVPR  = {IEEE Conf. Comput. Vis. Pattern Recog.})

@String(ICCV  = {Int. Conf. Comput. Vis.})

@String(ECCV  = {Eur. Conf. Comput. Vis.})

@String(CVPR  = {CVPR})

@String(ICCV  = {ICCV})

@String(ECCV  = {ECCV})

@String(CVPR= {IEEE Conf. Comput. Vis. Pattern Recog.})

@String(ICCV= {Int. Conf. Comput. Vis.})

@String(ECCV= {Eur. Conf. Comput. Vis.})

@inproceedings{ma2024vision,
   author = {Ma, Yecheng Jason and Hejna, Joey and Fu, Chuyuan and Shah, Dhruv and Liang, Jacky and Xu, Zhuo and Kirmani, Sean and Xu, Peng and Driess, Danny and Xiao, Ted and Bastani, Osbert and Jayaraman, Dinesh and Yu, Wenhao and Zhang, Tingnan and Sadigh, Dorsa and Xia, Fei},
 booktitle = {International Conference on Learning Representations},
 editor = {Y. Yue and A. Garg and N. Peng and F. Sha and R. Yu},
 pages = {33984--34009},
 title = {Vision Language Models are In-Context Value Learners},
 url = {https://proceedings.iclr.cc/paper_files/paper/2025/file/54854cf15a24fff9f5134a8641136fe4-Paper-Conference.pdf},
 volume = {2025},
 year = {2025}
}

@inproceedings{hudson2019gqa,
   author={Hudson, Drew A. and Manning, Christopher D.},
  booktitle={2019 IEEE/CVF Conference on Computer Vision and Pattern Recognition (CVPR)}, 
  title={GQA: A New Dataset for Real-World Visual Reasoning and Compositional Question Answering}, 
  year={2019},
  volume={},
  number={},
  pages={6693-6702},
  doi={10.1109/CVPR.2019.00686}
}

@inproceedings{wang2025spatial457,
  author={Wang, Xingrui and Ma, Wufei and Zhang, Tiezheng and de Melo, Celso M and Chen, Jieneng and Yuille, Alan},
  booktitle={2025 IEEE/CVF Conference on Computer Vision and Pattern Recognition (CVPR)}, 
  title={Spatial457: A Diagnostic Benchmark for 6D Spatial Reasoning of Large Multimodal Models}, 
  year={2025},
  volume={},
  number={},
  pages={24669-24679},
  doi={10.1109/CVPR52734.2025.02297}
}

@inproceedings{tong2024cambrian,
   author = {Tong, Shengbang and Brown, Ellis and Wu, Penghao and Woo, Sanghyun and Middepogu, Manoj and Akula, Sai Charitha and Yang, Jihan and Yang, Shusheng and Iyer, Adithya and Pan, Xichen and Wang, Austin and Fergus, Rob and LeCun, Yann and Xie, Saining},
 booktitle = {Advances in Neural Information Processing Systems},
 doi = {10.52202/079017-2771},
 editor = {A. Globerson and L. Mackey and D. Belgrave and A. Fan and U. Paquet and J. Tomczak and C. Zhang},
 pages = {87310--87356},
 publisher = {Curran Associates, Inc.},
 title = {Cambrian-1: A Fully Open, Vision-Centric Exploration of Multimodal LLMs},
 url = {https://proceedings.neurips.cc/paper_files/paper/2024/file/9ee3a664ccfeabc0da16ac6f1f1cfe59-Paper-Conference.pdf},
 volume = {37},
 year = {2024}
}

@inproceedings{zhou2025roborefer,
  author = {Zhou, Enshen and An, Jingkun and Chi, Cheng and Han, Yi and Rong, Shanyu and Zhang, Chi and Wang, Pengwei and Wang, Zhongyuan and Huang, Tiejun and Sheng, Lu and Zhang, Shanghang},
 booktitle = {Advances in Neural Information Processing Systems},
 doi = {10.52202/085713-0957},
 editor = {D. Belgrave and C. Zhang and H. Lin and R. Pascanu and P. Koniusz and M. Ghassemi and N. Chen},
 pages = {28404--28481},
 publisher = {Curran Associates, Inc.},
 title = {RoboRefer: Towards Spatial Referring with Reasoning in Vision-Language Models for Robotics},
 url = {https://proceedings.neurips.cc/paper_files/paper/2025/file/29416b66c2149872b9d1415a3fd2c5e0-Paper-Conference.pdf},
 volume = {38, Main Conference},
 year = {2025}
}

@inproceedings{song2025robospatial,
  author={Song, Chan Hee and Blukis, Valts and Tremblay, Jonathan and Tyree, Stephen and Su, Yu and Birchfield, Stan},
  booktitle={2025 IEEE/CVF Conference on Computer Vision and Pattern Recognition (CVPR)}, 
  title={RoboSpatial: Teaching Spatial Understanding to 2D and 3D Vision-Language Models for Robotics}, 
  year={2025},
  volume={},
  number={},
  pages={15768-15780},
  doi={10.1109/CVPR52734.2025.01470}
}

@article{yuan2024robopoint,
  title={Robopoint: A vision-language model for spatial affordance prediction for robotics},
  author={Yuan, Wentao and Duan, Jiafei and Blukis, Valts and Pumacay, Wilbert and Krishna, Ranjay and Murali, Adithyavairavan and Mousavian, Arsalan and Fox, Dieter},
  journal={arXiv preprint arXiv:2406.10721},
  year={2024}
}

@inproceedings{majumdar2024openeqa,
    author={Majumdar, Arjun and Ajay, Anurag and Zhang, Xiaohan and Putta, Pranav and Yenamandra, Sriram and Henaff, Mikael and Silwal, Sneha and Mcvay, Paul and Maksymets, Oleksandr and Arnaud, Sergio and Yadav, Karmesh and Li, Qiyang and Newman, Ben and Sharma, Mohit and Berges, Vincent and Zhang, Shiqi and Agrawal, Pulkit and Bisk, Yonatan and Batra, Dhruv and Kalakrishnan, Mrinal and Meier, Franziska and Paxton, Chris and Sax, Alexander and Rajeswaran, Aravind},
  booktitle={2024 IEEE/CVF Conference on Computer Vision and Pattern Recognition (CVPR)}, 
  title={OpenEQA: Embodied Question Answering in the Era of Foundation Models}, 
  year={2024},
  volume={},
  number={},
  pages={16488-16498},
  doi={10.1109/CVPR52733.2024.01560}
}

@inproceedings{yang2024think,
    author={Yang, Jihan and Yang, Shusheng and Gupta, Anjali W. and Han, Rilyn and Fei-Fei, Li and Xie, Saining},
  booktitle={2025 IEEE/CVF Conference on Computer Vision and Pattern Recognition (CVPR)}, 
  title={Thinking in Space: How Multimodal Large Language Models See, Remember, and Recall Spaces}, 
  year={2025},
  volume={},
  number={},
  pages={10632-10643},
  doi={10.1109/CVPR52734.2025.00994}
}

@inproceedings{fu2024blink,
author="Fu, Xingyu
and Hu, Yushi
and Li, Bangzheng
and Feng, Yu
and Wang, Haoyu
and Lin, Xudong
and Roth, Dan
and Smith, Noah A.
and Ma, Wei-Chiu
and Krishna, Ranjay",
editor="Leonardis, Ale{\v{s}}
and Ricci, Elisa
and Roth, Stefan
and Russakovsky, Olga
and Sattler, Torsten
and Varol, G{\"u}l",
title="BLINK: Multimodal Large Language Models Can See but Not Perceive",
booktitle="Computer Vision -- ECCV 2024",
year="2025",
publisher="Springer Nature Switzerland",
address="Cham",
pages="148--166",
isbn="978-3-031-73337-6"
}

@inproceedings{du2024embspatial,
    title = "{E}mb{S}patial-Bench: Benchmarking Spatial Understanding for Embodied Tasks with Large Vision-Language Models",
    author = "Du, Mengfei  and
      Wu, Binhao  and
      Li, Zejun  and
      Huang, Xuanjing  and
      Wei, Zhongyu",
    editor = "Ku, Lun-Wei  and
      Martins, Andre  and
      Srikumar, Vivek",
    booktitle = "Proceedings of the 62nd Annual Meeting of the Association for Computational Linguistics (Volume 2: Short Papers)",
    month = aug,
    year = "2024",
    address = "Bangkok, Thailand",
    publisher = "Association for Computational Linguistics",
    url = "https://aclanthology.org/2024.acl-short.33/",
    doi = "10.18653/v1/2024.acl-short.33",
    pages = "346--355"
}

@inproceedings{yin2025spatial,
  title={Spatial mental modeling from limited views},
  author={Yin, Baiqiao and Wang, Qineng and Zhang, Pingyue and Zhang, Jianshu and Wang, Kangrui and Wang, Zihan and Zhang, Jieyu and Chandrasegaran, Keshigeyan and Liu, Han and Krishna, Ranjay and Xie, Saining and Li, Manling and Wu, Jiajun and Fei-Fei, Li},
  booktitle={Structural Priors for Vision Workshop at ICCV'25},
  year={2025},
  url={https://openreview.net/forum?id=yVQVCAyVlT}
}

@article{ray2024sat,
  title={SAT: Dynamic Spatial Aptitude Training for Multimodal Language Models},
  author={Arijit Ray and Jiafei Duan and Ellis Brown and Reuben Tan and Dina Bashkirova and Rose Hendrix and Kiana Ehsani and Aniruddha Kembhavi and Bryan A. Plummer and Ranjay Krishna and Kuo-Hao Zeng and Kate Saenko},
  journal={arXiv preprint arXiv:2412.07755},
  year={2024}
}

@article{azzolini2025cosmos,
  title={Cosmos-Reason1: From Physical Common Sense To Embodied Reasoning}, 
  author={NVIDIA and Alisson Azzolini and Junjie Bai and Hannah Brandon and Jiaxin Cao and Prithvijit Chattopadhyay and Huayu Chen and Jinju Chu and Yin Cui and Jenna Diamond and Yifan Ding and Liang Feng and Francesco Ferroni and Rama Govindaraju and Jinwei Gu and Siddharth Gururani and Imad El Hanafi and Zekun Hao and Jacob Huffman and Jingyi Jin and Brendan Johnson and Rizwan Khan and George Kurian and Elena Lantz and Nayeon Lee and Zhaoshuo Li and Xuan Li and Maosheng Liao and Tsung-Yi Lin and Yen-Chen Lin and Ming-Yu Liu and Xiangyu Lu and Alice Luo and Andrew Mathau and Yun Ni and Lindsey Pavao and Wei Ping and David W. Romero and Misha Smelyanskiy and Shuran Song and Lyne Tchapmi and Andrew Z. Wang and Boxin Wang and Haoxiang Wang and Fangyin Wei and Jiashu Xu and Yao Xu and Dinghao Yang and Xiaodong Yang and Zhuolin Yang and Jingxu Zhang and Xiaohui Zeng and Zhe Zhang},
  year={2025},
  eprint={2503.15558},
  archivePrefix={arXiv},
  primaryClass={cs.AI},
  url={https://arxiv.org/abs/2503.15558}
}

@inproceedings{yang2025guiding,
   author={Yang, Zhutian and Garrett, Caelan and Fox, Dieter and Lozano-Pérez, Tomás and Kaelbling, Leslie Pack},
  booktitle={2025 IEEE International Conference on Robotics and Automation (ICRA)}, 
  title={Guiding Long-Horizon Task and Motion Planning with Vision Language Models}, 
  year={2025},
  volume={},
  number={},
  pages={16847-16853},
  doi={10.1109/ICRA55743.2025.11128705}
}

@inproceedings{hancock2025actions,
   author = {Hancock, Asher and Wu, Xindi and Zha, Lihan and Russakovsky, Olga and Majumdar, Anirudha},
 booktitle = {International Conference on Learning Representations},
 editor = {C. Vondrick and B. Hariharan and C. Raffel and L. Pinto and D. Yang and A. Faust},
 pages = {75333--75360},
 title = {Actions as Language: Fine-Tuning VLMs into VLAs Without Catastrophic Forgetting},
 url = {https://proceedings.iclr.cc/paper_files/paper/2026/file/7a0f8055c838df8e62329a76c7c6403d-Paper-Conference.pdf},
 volume = {2026},
 year = {2026}
}

@inproceedings{liang2022code,
  author={Liang, Jacky and Huang, Wenlong and Xia, Fei and Xu, Peng and Hausman, Karol and Ichter, Brian and Florence, Pete and Zeng, Andy},
  booktitle={2023 IEEE International Conference on Robotics and Automation (ICRA)}, 
  title={Code as Policies: Language Model Programs for Embodied Control}, 
  year={2023},
  volume={},
  number={},
  pages={9493-9500},
  doi={10.1109/ICRA48891.2023.10160591}
}

@inproceedings{rocamonde2023vision,
   author = {Rocamonde, Juan and Montesinos, Victoriano and Nava, Elvis and Perez, Ethan and Lindner, David},
 booktitle = {International Conference on Learning Representations},
 editor = {B. Kim and Y. Yue and S. Chaudhuri and K. Fragkiadaki and M. Khan and Y. Sun},
 pages = {28446--28463},
 title = {Vision-Language Models are Zero-Shot Reward Models for Reinforcement Learning},
 url = {https://proceedings.iclr.cc/paper_files/paper/2024/file/7a7f6cc5dc2a84fb4edf0feb8e5cfd50-Paper-Conference.pdf},
 volume = {2024},
 year = {2024}
}

@article{chen2025robo2vlm,
  title={Robo2vlm: Visual question answering from large-scale in-the-wild robot manipulation datasets},
  author={Chen, Kaiyuan and Xie, Shuangyu and Ma, Zehan and Sanketi, Pannag R and Goldberg, Ken},
  journal={arXiv preprint arXiv:2505.15517},
  year={2025}
}

@inproceedings{balepur2024artifacts,
    title = "Artifacts or Abduction: How Do {LLM}s Answer Multiple-Choice Questions Without the Question?",
    author = "Balepur, Nishant  and
      Ravichander, Abhilasha  and
      Rudinger, Rachel",
    editor = "Ku, Lun-Wei  and
      Martins, Andre  and
      Srikumar, Vivek",
    booktitle = "Proceedings of the 62nd Annual Meeting of the Association for Computational Linguistics (Volume 1: Long Papers)",
    month = aug,
    year = "2024",
    address = "Bangkok, Thailand",
    publisher = "Association for Computational Linguistics",
    url = "https://aclanthology.org/2024.acl-long.555/",
    doi = "10.18653/v1/2024.acl-long.555",
    pages = "10308--10330"
}

@inproceedings{zheng2023large,
  author = {Zheng, Chujie and Zhou, Hao and Meng, Fandong and Zhou, Jie and Huang, Minlie},
 booktitle = {International Conference on Learning Representations},
 editor = {B. Kim and Y. Yue and S. Chaudhuri and K. Fragkiadaki and M. Khan and Y. Sun},
 pages = {19426--19454},
 title = {Large Language Models Are Not Robust Multiple Choice Selectors},
 url = {https://proceedings.iclr.cc/paper_files/paper/2024/file/54dd9e0cff6d9214e20d97eb2a3bae49-Paper-Conference.pdf},
 volume = {2024},
 year = {2024}
}

@article{khazatsky2024droid,
  title={Droid: A large-scale in-the-wild robot manipulation dataset},
  author={Alexander Khazatsky and Karl Pertsch and Suraj Nair and Ashwin Balakrishna and Sudeep Dasari and Siddharth Karamcheti and Soroush Nasiriany and Mohan Kumar Srirama and Lawrence Yunliang Chen and Kirsty Ellis and Peter David Fagan and Joey Hejna and Masha Itkina and Marion Lepert and Yecheng Jason Ma and Patrick Tree Miller and Jimmy Wu and Suneel Belkhale and Shivin Dass and Huy Ha and Arhan Jain and Abraham Lee and Youngwoon Lee and Marius Memmel and Sungjae Park and Ilija Radosavovic and Kaiyuan Wang and Albert Zhan and Kevin Black and Cheng Chi and Kyle Beltran Hatch and Shan Lin and Jingpei Lu and Jean Mercat and Abdul Rehman and Pannag R Sanketi and Archit Sharma and Cody Simpson and Quan Vuong and Homer Rich Walke and Blake Wulfe and Ted Xiao and Jonathan Heewon Yang and Arefeh Yavary and Tony Z. Zhao and Christopher Agia and Rohan Baijal and Mateo Guaman Castro and Daphne Chen and Qiuyu Chen and Trinity Chung and Jaimyn Drake and Ethan Paul Foster and Jensen Gao and Vitor Guizilini and David Antonio Herrera and Minho Heo and Kyle Hsu and Jiaheng Hu and Muhammad Zubair Irshad and Donovon Jackson and Charlotte Le and Yunshuang Li and Kevin Lin and Roy Lin and Zehan Ma and Abhiram Maddukuri and Suvir Mirchandani and Daniel Morton and Tony Nguyen and Abigail O'Neill and Rosario Scalise and Derick Seale and Victor Son and Stephen Tian and Emi Tran and Andrew E. Wang and Yilin Wu and Annie Xie and Jingyun Yang and Patrick Yin and Yunchu Zhang and Osbert Bastani and Glen Berseth and Jeannette Bohg and Ken Goldberg and Abhinav Gupta and Abhishek Gupta and Dinesh Jayaraman and Joseph J Lim and Jitendra Malik and Roberto Martín-Martín and Subramanian Ramamoorthy and Dorsa Sadigh and Shuran Song and Jiajun Wu and Michael C. Yip and Yuke Zhu and Thomas Kollar and Sergey Levine and Chelsea Finn},
  journal={arXiv preprint arXiv:2403.12945},
  year={2024}
}

@INPROCEEDINGS{10611477,
  author={O’Neill, Abby and Rehman, Abdul and Maddukuri, Abhiram and Gupta, Abhishek and Padalkar, Abhishek and Lee, Abraham and Pooley, Acorn and Gupta, Agrim and Mandlekar, Ajay and Jain, Ajinkya and Tung, Albert and Bewley, Alex and Herzog, Alex and Irpan, Alex and Khazatsky, Alexander and Rai, Anant and Gupta, Anchit and Wang, Andrew and Singh, Anikait and Garg, Animesh and Kembhavi, Aniruddha and Xie, Annie and Brohan, Anthony and Raffin, Antonin and Sharma, Archit and Yavary, Arefeh and Jain, Arhan and Balakrishna, Ashwin and Wahid, Ayzaan and Burgess-Limerick, Ben and Kim, Beomjoon and Schölkopf, Bernhard and Wulfe, Blake and Ichter, Brian and Lu, Cewu and Xu, Charles and Le, Charlotte and Finn, Chelsea and Wang, Chen and Xu, Chenfeng and Chi, Cheng and Huang, Chenguang and Chan, Christine and Agia, Christopher and Pan, Chuer and Fu, Chuyuan and Devin, Coline and Xu, Danfei and Morton, Daniel and Driess, Danny and Chen, Daphne and Pathak, Deepak and Shah, Dhruv and Büchler, Dieter and Jayaraman, Dinesh and Kalashnikov, Dmitry and Sadigh, Dorsa and Johns, Edward and Foster, Ethan and Liu, Fangchen and Ceola, Federico and Xia, Fei and Zhao, Feiyu and Stulp, Freek and Zhou, Gaoyue and Sukhatme, Gaurav S. and Salhotra, Gautam and Yan, Ge and Feng, Gilbert and Schiavi, Giulio and Berseth, Glen and Kahn, Gregory and Wang, Guanzhi and Su, Hao and Fang, Hao-Shu and Shi, Haochen and Bao, Henghui and Ben Amor, Heni and Christensen, Henrik I and Furuta, Hiroki and Walke, Homer and Fang, Hongjie and Ha, Huy and Mordatch, Igor and Radosavovic, Ilija and Leal, Isabel and Liang, Jacky and Abou-Chakra, Jad and Kim, Jaehyung and Drake, Jaimyn and Peters, Jan and Schneider, Jan and Hsu, Jasmine and Bohg, Jeannette and Bingham, Jeffrey and Wu, Jeffrey and Gao, Jensen and Hu, Jiaheng and Wu, Jiajun and Wu, Jialin and Sun, Jiankai and Luo, Jianlan and Gu, Jiayuan and Tan, Jie and Oh, Jihoon and Wu, Jimmy and Lu, Jingpei and Yang, Jingyun and Malik, Jitendra and Silvério, João and Hejna, Joey and Booher, Jonathan and Tompson, Jonathan and Yang, Jonathan and Salvador, Jordi and Lim, Joseph J. and Han, Junhyek and Wang, Kaiyuan and Rao, Kanishka and Pertsch, Karl and Hausman, Karol and Go, Keegan and Gopalakrishnan, Keerthana and Goldberg, Ken and Byrne, Kendra and Oslund, Kenneth and Kawaharazuka, Kento and Black, Kevin and Lin, Kevin and Zhang, Kevin and Ehsani, Kiana and Lekkala, Kiran and Ellis, Kirsty and Rana, Krishan and Srinivasan, Krishnan and Fang, Kuan and Singh, Kunal Pratap and Zeng, Kuo-Hao and Hatch, Kyle and Hsu, Kyle and Itti, Laurent and Chen, Lawrence Yunliang and Pinto, Lerrel and Fei-Fei, Li and Tan, Liam and Fan, Linxi Jim and Ott, Lionel and Lee, Lisa and Weihs, Luca and Chen, Magnum and Lepert, Marion and Memmel, Marius and Tomizuka, Masayoshi and Itkina, Masha and Castro, Mateo Guaman and Spero, Max and Du, Maximilian and Ahn, Michael and Yip, Michael C. and Zhang, Mingtong and Ding, Mingyu and Heo, Minho and Srirama, Mohan Kumar and Sharma, Mohit and Kim, Moo Jin and Kanazawa, Naoaki and Hansen, Nicklas and Heess, Nicolas and Joshi, Nikhil J and Suenderhauf, Niko and Liu, Ning and Di Palo, Norman and Shafiullah, Nur Muhammad Mahi and Mees, Oier and Kroemer, Oliver and Bastani, Osbert and Sanketi, Pannag R and Miller, Patrick Tree and Yin, Patrick and Wohlhart, Paul and Xu, Peng and Fagan, Peter David and Mitrano, Peter and Sermanet, Pierre and Abbeel, Pieter and Sundaresan, Priya and Chen, Qiuyu and Vuong, Quan and Rafailov, Rafael and Tian, Ran and Doshi, Ria and Martín-Martín, Roberto and Baijal, Rohan and Scalise, Rosario and Hendrix, Rose and Lin, Roy and Qian, Runjia and Zhang, Ruohan and Mendonca, Russell and Shah, Rutav and Hoque, Ryan and Julian, Ryan and Bustamante, Samuel and Kirmani, Sean and Levine, Sergey and Lin, Shan and Moore, Sherry and Bahl, Shikhar and Dass, Shivin and Sonawani, Shubham and Song, Shuran and Xu, Sichun and Haldar, Siddhant and Karamcheti, Siddharth and Adebola, Simeon and Guist, Simon and Nasiriany, Soroush and Schaal, Stefan and Welker, Stefan and Tian, Stephen and Ramamoorthy, Subramanian and Dasari, Sudeep and Belkhale, Suneel and Park, Sungjae and Nair, Suraj and Mirchandani, Suvir and Osa, Takayuki and Gupta, Tanmay and Harada, Tatsuya and Matsushima, Tatsuya and Xiao, Ted and Kollar, Thomas and Yu, Tianhe and Ding, Tianli and Davchev, Todor and Zhao, Tony Z. and Armstrong, Travis and Darrell, Trevor and Chung, Trinity and Jain, Vidhi and Vanhoucke, Vincent and Zhan, Wei and Zhou, Wenxuan and Burgard, Wolfram and Chen, Xi and Wang, Xiaolong and Zhu, Xinghao and Geng, Xinyang and Liu, Xiyuan and Liangwei, Xu and Li, Xuanlin and Lu, Yao and Ma, Yecheng Jason and Kim, Yejin and Chebotar, Yevgen and Zhou, Yifan and Zhu, Yifeng and Wu, Yilin and Xu, Ying and Wang, Yixuan and Bisk, Yonatan and Cho, Yoonyoung and Lee, Youngwoon and Cui, Yuchen and Cao, Yue and Wu, Yueh-Hua and Tang, Yujin and Zhu, Yuke and Zhang, Yunchu and Jiang, Yunfan and Li, Yunshuang and Li, Yunzhu and Iwasawa, Yusuke and Matsuo, Yutaka and Ma, Zehan and Xu, Zhuo and Cui, Zichen Jeff and Zhang, Zichen and Lin, Zipeng},
  booktitle={2024 IEEE International Conference on Robotics and Automation (ICRA)}, 
  title={Open X-Embodiment: Robotic Learning Datasets and RT-X Models : Open X-Embodiment Collaboration0}, 
  year={2024},
  volume={},
  number={},
  pages={6892-6903},
  doi={10.1109/ICRA57147.2024.10611477}}

@inproceedings{antol2015vqa,
  title={Vqa: Visual question answering},
  author={Antol, Stanislaw and Agrawal, Aishwarya and Lu, Jiasen and Mitchell, Margaret and Batra, Dhruv and Zitnick, C Lawrence and Parikh, Devi},
  booktitle={Proceedings of the IEEE international conference on computer vision},
  pages={2425--2433},
  year={2015}
}

@article{chang2017matterport3d,
  title={Matterport3d: Learning from rgb-d data in indoor environments},
  author={Chang, Angel and Dai, Angela and Funkhouser, Thomas and Halber, Maciej and Niessner, Matthias and Savva, Manolis and Song, Shuran and Zeng, Andy and Zhang, Yinda},
  journal={arXiv preprint arXiv:1709.06158},
  year={2017}
}

@article{kolve2017ai2,
  title={Ai2-thor: An interactive 3d environment for visual ai},
  author={Eric Kolve and Roozbeh Mottaghi and Winson Han and Eli VanderBilt and Luca Weihs and Alvaro Herrasti and Matt Deitke and Kiana Ehsani and Daniel Gordon and Yuke Zhu and Aniruddha Kembhavi and Abhinav Gupta and Ali Farhadi},
  journal={arXiv preprint arXiv:1712.05474},
  year={2017}
}

@inproceedings{deitke2022,
 author = {Deitke, Matt and VanderBilt, Eli and Herrasti, Alvaro and Weihs, Luca and Ehsani, Kiana and Salvador, Jordi and Han, Winson and Kolve, Eric and Kembhavi, Aniruddha and Mottaghi, Roozbeh},
 booktitle = {Advances in Neural Information Processing Systems},
 doi = {10.52202/068431-0433},
 editor = {S. Koyejo and S. Mohamed and A. Agarwal and D. Belgrave and K. Cho and A. Oh},
 pages = {5982--5994},
 publisher = {Curran Associates, Inc.},
 title = {{ProcTHOR}: Large-Scale Embodied AI Using Procedural Generation},
 url = {https://proceedings.neurips.cc/paper_files/paper/2022/file/27c546ab1e4f1d7d638e6a8dfbad9a07-Paper-Conference.pdf},
 volume = {35},
 year = {2022}
}

@inproceedings{dai2017scannet,
  author={Dai, Angela and Chang, Angel X. and Savva, Manolis and Halber, Maciej and Funkhouser, Thomas and Nießner, Matthias},
  booktitle={2017 IEEE Conference on Computer Vision and Pattern Recognition (CVPR)}, 
  title={ScanNet: Richly-Annotated 3D Reconstructions of Indoor Scenes}, 
  year={2017},
  volume={},
  number={},
  pages={2432-2443},
  doi={10.1109/CVPR.2017.261}
}

@inproceedings{fang2020graspnet,
  author={Fang, Hao-Shu and Wang, Chenxi and Gou, Minghao and Lu, Cewu},
  booktitle={2020 IEEE/CVF Conference on Computer Vision and Pattern Recognition (CVPR)}, 
  title={GraspNet-1Billion: A Large-Scale Benchmark for General Object Grasping}, 
  year={2020},
  volume={},
  number={},
  pages={11441-11450},
  doi={10.1109/CVPR42600.2020.01146}
}

@inproceedings{tyree20226,
  author={Tyree, Stephen and Tremblay, Jonathan and To, Thang and Cheng, Jia and Mosier, Terry and Smith, Jeffrey and Birchfield, Stan},
  booktitle={2022 IEEE/RSJ International Conference on Intelligent Robots and Systems (IROS)}, 
  title={{6-DoF} Pose Estimation of Household Objects for Robotic Manipulation: An Accessible Dataset and Benchmark}, 
  year={2022},
  volume={},
  number={},
  pages={13081-13088},
  doi={10.1109/IROS47612.2022.9981838}
}

@article{li2025otter,
  author={Li, Bo and Zhang, Yuanhan and Chen, Liangyu and Wang, Jinghao and Pu, Fanyi and Cahyono, Joshua Adrian and Yang, Jingkang and Li, Chunyuan and Liu, Ziwei},
  journal={IEEE Transactions on Pattern Analysis and Machine Intelligence}, 
  title={Otter: A Multi-Modal Model With In-Context Instruction Tuning}, 
  year={2025},
  volume={47},
  number={9},
  pages={7543-7557},
  doi={10.1109/TPAMI.2025.3571946}
}

@inproceedings{liu2024grounding,
  author="Liu, Shilong
and Zeng, Zhaoyang
and Ren, Tianhe
and Li, Feng
and Zhang, Hao
and Yang, Jie
and Jiang, Qing
and Li, Chunyuan
and Yang, Jianwei
and Su, Hang
and Zhu, Jun
and Zhang, Lei",
editor="Leonardis, Ale{\v{s}}
and Ricci, Elisa
and Roth, Stefan
and Russakovsky, Olga
and Sattler, Torsten
and Varol, G{\"u}l",
title="Grounding DINO: Marrying DINO with Grounded Pre-training for Open-Set Object Detection",
booktitle="Computer Vision -- ECCV 2024",
year="2025",
publisher="Springer Nature Switzerland",
address="Cham",
pages="38--55",
isbn="978-3-031-72970-6"
}

@article{ke2025marigold,
  author={Ke, Bingxin and Qu, Kevin and Wang, Tianfu and Metzger, Nando and Huang, Shengyu and Li, Bo and Obukhov, Anton and Schindler, Konrad},
  journal={IEEE Transactions on Pattern Analysis and Machine Intelligence}, 
  title={Marigold: Affordable Adaptation of Diffusion-Based Image Generators for Image Analysis}, 
  year={2025},
  volume={},
  number={},
  pages={1-18},
  doi={10.1109/TPAMI.2025.3591076}
}

@misc{ren2024grounded,
      title={Grounded SAM: Assembling Open-World Models for Diverse Visual Tasks}, 
      author={Tianhe Ren and Shilong Liu and Ailing Zeng and Jing Lin and Kunchang Li and He Cao and Jiayu Chen and Xinyu Huang and Yukang Chen and Feng Yan and Zhaoyang Zeng and Hao Zhang and Feng Li and Jie Yang and Hongyang Li and Qing Jiang and Lei Zhang},
      year={2024},
      eprint={2401.14159},
      archivePrefix={arXiv},
      primaryClass={cs.CV}
}

@misc{gemini2025geminirobotics15,
      title={Gemini Robotics 1.5: Pushing the Frontier of Generalist Robots with Advanced Embodied Reasoning, Thinking, and Motion Transfer}, 
      author={Gemini Robotics Team and Abbas Abdolmaleki and Saminda Abeyruwan and Joshua Ainslie and Jean-Baptiste Alayrac and Montserrat Gonzalez Arenas and Ashwin Balakrishna and Nathan Batchelor and Alex Bewley and Jeff Bingham and Michael Bloesch and Konstantinos Bousmalis and Philemon Brakel and Anthony Brohan and Thomas Buschmann and Arunkumar Byravan and Serkan Cabi and Ken Caluwaerts and Federico Casarini and Christine Chan and Oscar Chang and London Chappellet-Volpini and Jose Enrique Chen and Xi Chen and Hao-Tien Lewis Chiang and Krzysztof Choromanski and Adrian Collister and David B. D'Ambrosio and Sudeep Dasari and Todor Davchev and Meet Kirankumar Dave and Coline Devin and Norman Di Palo and Tianli Ding and Carl Doersch and Adil Dostmohamed and Yilun Du and Debidatta Dwibedi and Sathish Thoppay Egambaram and Michael Elabd and Tom Erez and Xiaolin Fang and Claudio Fantacci and Cody Fong and Erik Frey and Chuyuan Fu and Ruiqi Gao and Marissa Giustina and Keerthana Gopalakrishnan and Laura Graesser and Oliver Groth and Agrim Gupta and Roland Hafner and Steven Hansen and Leonard Hasenclever and Sam Haves and Nicolas Heess and Brandon Hernaez and Alex Hofer and Jasmine Hsu and Lu Huang and Sandy H. Huang and Atil Iscen and Mithun George Jacob and Deepali Jain and Sally Jesmonth and Abhishek Jindal and Ryan Julian and Dmitry Kalashnikov and M. Emre Karagozler and Stefani Karp and Matija Kecman and J. Chase Kew and Donnie Kim and Frank Kim and Junkyung Kim and Thomas Kipf and Sean Kirmani and Ksenia Konyushkova and Li Yang Ku and Yuheng Kuang and Thomas Lampe and Antoine Laurens and Tuan Anh Le and Isabel Leal and Alex X. Lee and Tsang-Wei Edward Lee and Guy Lever and Jacky Liang and Li-Heng Lin and Fangchen Liu and Shangbang Long and Caden Lu and Sharath Maddineni and Anirudha Majumdar and Kevis-Kokitsi Maninis and Andrew Marmon and Sergio Martinez and Assaf Hurwitz Michaely and Niko Milonopoulos and Joss Moore and Robert Moreno and Michael Neunert and Francesco Nori and Joy Ortiz and Kenneth Oslund and Carolina Parada and Emilio Parisotto and Amaris Paryag and Acorn Pooley and Thomas Power and Alessio Quaglino and Haroon Qureshi and Rajkumar Vasudeva Raju and Helen Ran and Dushyant Rao and Kanishka Rao and Isaac Reid and David Rendleman and Krista Reymann and Miguel Rivas and Francesco Romano and Yulia Rubanova and Peter Pastor Sampedro and Pannag R Sanketi and Dhruv Shah and Mohit Sharma and Kathryn Shea and Mohit Shridhar and Charles Shu and Vikas Sindhwani and Sumeet Singh and Radu Soricut and Rachel Sterneck and Ian Storz and Razvan Surdulescu and Jie Tan and Jonathan Tompson and Saran Tunyasuvunakool and Jake Varley and Grace Vesom and Giulia Vezzani and Maria Bauza Villalonga and Oriol Vinyals and René Wagner and Ayzaan Wahid and Stefan Welker and Paul Wohlhart and Chengda Wu and Markus Wulfmeier and Fei Xia and Ted Xiao and Annie Xie and Jinyu Xie and Peng Xu and Sichun Xu and Ying Xu and Zhuo Xu and Jimmy Yan and Sherry Yang and Skye Yang and Yuxiang Yang and Hiu Hong Yu and Wenhao Yu and Wentao Yuan and Yuan Yuan and Jingwei Zhang and Tingnan Zhang and Zhiyuan Zhang and Allan Zhou and Guangyao Zhou and Yuxiang Zhou},
      year={2025},
      eprint={2510.03342},
      archivePrefix={arXiv},
      primaryClass={cs.RO},
      url={https://arxiv.org/abs/2510.03342}, 
}

@article{shao2024deepseekmath,
  title={Deepseekmath: Pushing the limits of mathematical reasoning in open language models},
  author={Zhihong Shao and Peiyi Wang and Qihao Zhu and Runxin Xu and Junxiao Song and Xiao Bi and Haowei Zhang and Mingchuan Zhang and Y. K. Li and Y. Wu and Daya Guo},
  journal={arXiv preprint arXiv:2402.03300},
  year={2024}
}

@article{ye2025robofac,
  title={Robofac: A comprehensive framework for robotic failure analysis and correction},
  author={Ye, Zewei and Lu, Weifeng and Ye, Minghao and Lin, Tao and Yang, Shuo and Yan, Junchi and Zhao, Bo},
  journal={arXiv preprint arXiv:2505.12224},
  year={2025}
}

@article{brohan2022rt,
  title={Rt-1: Robotics transformer for real-world control at scale},
  author={Anthony Brohan and Noah Brown and Justice Carbajal and Yevgen Chebotar and Joseph Dabis and Chelsea Finn and Keerthana Gopalakrishnan and Karol Hausman and Alex Herzog and Jasmine Hsu and Julian Ibarz and Brian Ichter and Alex Irpan and Tomas Jackson and Sally Jesmonth and Nikhil J Joshi and Ryan Julian and Dmitry Kalashnikov and Yuheng Kuang and Isabel Leal and Kuang-Huei Lee and Sergey Levine and Yao Lu and Utsav Malla and Deeksha Manjunath and Igor Mordatch and Ofir Nachum and Carolina Parada and Jodilyn Peralta and Emily Perez and Karl Pertsch and Jornell Quiambao and Kanishka Rao and Michael Ryoo and Grecia Salazar and Pannag Sanketi and Kevin Sayed and Jaspiar Singh and Sumedh Sontakke and Austin Stone and Clayton Tan and Huong Tran and Vincent Vanhoucke and Steve Vega and Quan Vuong and Fei Xia and Ted Xiao and Peng Xu and Sichun Xu and Tianhe Yu and Brianna Zitkovich},
  journal={arXiv preprint arXiv:2212.06817},
  year={2022}
}

@article{shafiullah2023bringing,
  title={On bringing robots home},
  author={Shafiullah, Nur Muhammad Mahi and Rai, Anant and Etukuru, Haritheja and Liu, Yiqian and Misra, Ishan and Chintala, Soumith and Pinto, Lerrel},
  journal={arXiv preprint arXiv:2311.16098},
  year={2023}
}

@article{xia2026roboprocessbench,
title={RoboProcessBench: Benchmarking Process-Aware Understanding in Vision-Language Robotic Manipulation},
  author={Xia, Dayu and Shi, Yue and Mu, Yao and Ji, Huiting and Ma, Chaofan and Zhou, Yingjie and Chen, Hua and Liu, Yang and Cao, Jiezhang and Zhai, Guangtao},
  journal={arXiv preprint arXiv:2606.13040},
  year={2026}
}

@inproceedings{chuan2024can,
 author = {Wen, Chuan and Jayaraman, Dinesh and Gao, Yang},
 booktitle = {International Conference on Learning Representations},
 editor = {B. Kim and Y. Yue and S. Chaudhuri and K. Fragkiadaki and M. Khan and Y. Sun},
 pages = {35792--35812},
 title = {Can Transformers Capture Spatial Relations between Objects?},
 url = {https://proceedings.iclr.cc/paper_files/paper/2024/file/9973907a2acfd04f030cd244e6a0fe35-Paper-Conference.pdf},
 volume = {2024},
 year = {2024}
}

\newpage
\appendix
\section{Related Works}
\label{app:rel_works}
\textbf{Robot Manipulation Datasets}. Large scale robotics datasets such as Open X-Embodiment \cite{10611477} and DROID \cite{khazatsky2024droid} contain robot manipulation trajectory data alongside associated images captured from cameras attached to the robot platform. These have been repurposed into evaluating VLMs for auto-regressive value estimation \cite{ma2024vision} and visual question-answering \cite{chen2025robo2vlm}. While VLM spatial understanding has been explored using images from DROID \cite{chen2025robo2vlm}, such tasks have focused on the spatial orientation and the interactions between the robot gripper and the object it interacts with in that particular trajectory, with ground truths sourced from the underlying robot gripper data. Existing benchmarks do not evaluate VLMs on their spatiotemporal understanding of the entire scene as viewed from the perspective of the robot, and we aim to address some of these limitations by analyzing the images associated with the trajectory to construct a spatiotemporal ordering task.

\textbf{Visual Question Answering Tasks}. Both open-ended and MCQ-based visual question answering benchmarks have long been favored for evaluating VLMs on 2D spatial reasoning \cite{hudson2019gqa, antol2015vqa, chuan2024can}). Such benchmarks have long focused on relative spatial orientation in 2D image space, and recently, this approach has been extended 3D reasoning by several benchmarks that ask VLMs to reason about how close an object is to the camera \cite{wang2025spatial457, tong2024cambrian, zhou2025roborefer, du2024embspatial, fu2024blink, ray2024sat}. 

However, such benchmarks suffer from two primary limitations. Firstly, they often ask questions about the relative depth of pairs of objects, which allows the model to guess its answer from two presented options or exhibit selection bias \cite{zheng2023large, balepur2024artifacts}.  
To mitigate this issue, some existing VLM reasoning benchmarks have examined evaluating VLMs using pointing-based questions \cite{yuan2024robopoint, song2025robospatial}. However, framing reasoning questions as pointing questions requires models to both solve the task at hand and learn to point reliably to the solution in image space, making it difficult to isolate VLM spatiotemporal reasoning capabilities.
Secondly, several of these benchmarks primarily use images from large Internet-sourced image datasets that are unlikely for present-day robots to observe in a real-world robot manipulation setting. While some embodied reasoning benchmarks \cite{majumdar2024openeqa, zhou2025roborefer, du2024embspatial, ray2024sat} use images and environments from indoor settings such as Matterport3D \cite{chang2017matterport3d}, AI2-Thor \cite{kolve2017ai2}, ProcThor \cite{deitke2022} and ScanNet \cite{dai2017scannet}, as well as tabletop datasets like GraspNET-1B \cite{fang2020graspnet} and HOPE \cite{tyree20226}. More recent benchmarks, however, have noted that due to the differences in properties like resolution, lighting variations and real-world robot, as well as the sim-to-real gap, it is difficult to generalize from these datasets to images captured during real-world robot manipulation \cite{chen2025robo2vlm, song2025robospatial}. We handle these considerations by ensuring that our tasks use images associated with real-world robot manipulation trajectories in the DROID dataset \cite{khazatsky2024droid} and are formatted as spatiotemporal ordering tasks with a combinatorial number of possible answers. 

\textbf{Spatiotemporal Reasoning Benchmarks}: Other approaches for evaluating spatial reasoning in multi-modal LLMs have focused on reasoning from multiple limited views \cite{yin2025spatial}, multiple perspectives \cite{azzolini2025cosmos}, and object localization tasks \cite{majumdar2024openeqa}. Approaches to temporal ordering have both focused on ordering video frames \cite{azzolini2025cosmos}
and reasoning about the order of occurrence of events in videos \cite{yang2024think, li2025otter}. However, such approaches have not evaluated VLMs on reasoning about real-world robot manipulation videos and images, due to the engineering difficulties associated with retrieving accurate ground-truth data for such questions from low-resolution robot camera images, which has only recently started to be explored in concurrent works to ours \cite{xia2026roboprocessbench}. Our approach instead focuses on spatiotemporal ordering tasks with ground truth information that is easily retrievable from the trajectory video or using traditional computer vision models. 

\section{Downstream Task Evaluations}
\label{app:large_scale_training_task}

To evaluate reasoning performance of models finetuned on a large RoboReason-generated dataset, we train Qwen3-VL-8B using SFT and GRPO on 8.8K task samples of each type. Training was conducted on three systems: one comprising a single NVIDIA B300 GPU with 262 GB VRAM for 36 hours, one comprising 4 NVIDIA H200 GPUs with 564GB VRAM for 15 hours, and one comprising 8 NVIDIA RTX A6000 GPUs for 3 hours. For temporal checkpoints, we train on the 3 image temporal prediction task. For depth and left-to-right checkpoints, we train on $k$-object ordering, where $k$ represents all valid objects and labels that remain after the respective filtering pipelines. The training data was constructed with seeds $s=12345$ for the temporal task, $s=54321$ for the depth task, and $s=2025$ for the left-to-right task. While training the 8B models using SFT and GRPO, the seed $s=42$ was used.

For evaluating downstream reasoning, we evaluate on 1) 150 question-answer pairs in the SAT test set \cite{ray2024sat}, 2) 1.9K multiple-choice and 5.9K open-ended robotics failure reasoning questions (with accuracy measured in terms of exact string match for open-ended questions) in the RoboFAC dataset \cite{ye2025robofac}, and 3) value estimation tasks on 1000 trajectories each from RT-1 \cite{brohan2022rt} and Dobb-E \cite{shafiullah2023bringing}, using the value estimation task format proposed in Generative Value Learning (GVL) \cite{ma2024vision}. For SAT and RoboFAC, we report task results by measuring the percentage of questions that were answered correctly. For the GVL tasks, we report the absolute average Value-Order Correlation (VOC), where the VOC of an ordering is the rank correlation between predicted values and the original chronological order \cite{ma2024vision}. For instances where the VLM did not successfully produce a ranking, we assume VOC to be 0.

\section{Impact of Camera View on Temporal Reasoning }
\label{app:camera_view_impact}
\begin{table}[H]
\tiny
\centering
\begin{tabular}{ccccc}
\toprule 
Model & Left & Right & Wrist & Multi-
\\
Name & View & View & Camera & View \\
\midrule
GPT-5 & \textbf{50.67} & 46.67 & 47.67 & 47.33 \\
GPT-5-Nano & \textbf{26.33} & 21.00 & 22.33 & 23.67 \\
Gemini-ER-1.5 & 42.33 & 42.00 & \textbf{45.00} & 38.00 \\
Qwen3-VL-235B & \textbf{36.00} & 28.67 & 35.33 & 32.67\\
\bottomrule
\end{tabular}
\caption{High Reasoning Model Accuracies (\%) for Single-View Image in 3 Image Base Setting for Temporal Ordering}
\label{tab:single_view}
\end{table}

In the temporal ordering task, unlike the spatial reasoning tasks, we provide the model with multiple camera views in the canonical setting. This is primarily because unlike the depth and left-to-right ordering tasks, minute differences between frames are often helpful for human evaluation of the temporal task. In the DROID setup, these differences may sometimes be occluded in specific camera views, necessitating the provision of all 3 camera views.

To evaluate if VLMs perform better when provided with one of the camera views, we run an ablation providing only the left, right, and wrist camera view to the VLM (with the high reasoning setting) in the 3 image ordering version of the task. Since this is simply the base setting of the task, no images had any overlays on them. We present our results in Table \ref{tab:single_view}. For all models and all views (except GPT-5-Nano and Qwen3-VL-235B while using the right-view image), using a single view for temporal ordering performs better than or comparable to (i.e. with a difference in performance smaller than 2\%) using an image composited with multiple camera views. This suggests that VLMs struggle to synthesize information from multiple views to reason about the scene in the temporal ordering task. This behavior is hence consistent with results on benchmarks like MindCube \cite{yin2025spatial}, which show that VLMs find it difficult to answer spatial understanding questions requiring information spread across multiple views.

\section{Spatiotemporal Ordering Task Construction and Prompts}

We design 3 spatiotemporal ordering tasks for evaluating VLM performance. The 2B model finetuning experiments and evaluations are conducted on a machine with 2× Intel(R) Xeon(R) Gold 6342 CPUs (96 cores total) and 8 NVIDIA RTX A6000 GPUs. The underlying robot trajectory images are taken from the DROID dataset \cite{khazatsky2024droid}, which has released its data under a CC-BY 4.0 license. We evaluate different input image types on them as follows: 

\begin{itemize}[leftmargin=1.25em]
    \item \textbf{Temporal Ordering}: Base Images, Segmentation Map Overlay and Bounding Box mask overlay (in 3-image and 5-image settings across low, medium and high reasoning levels)
    \item \textbf{Depth Ordering}: Base Image, Segmentation Map Overlay and Bounding Box mask overlay (in 3-object and 5-object settings across low, medium and high reasoning levels)
    \item \textbf{Left to Right Ordering}: Base Image, Segmentation Map Overlay and Bounding Box mask overlay (in 3-object and 5-object settings across low, medium and high reasoning levels)
\end{itemize}

For each ordering task, we also evaluate models on 2 and 4 image/object settings using the base images across low, medium, and high reasoning levels. Additionally, we also examine if structuring test-time compute helps improve VLM performance by evaluating models in the high reasoning setting using merge sort.

\subsection{Temporal Ordering Task}

Here the model sees a shuffled sequence of \emph{multiple} images from a robot task sequence and must decide which came first, second, etc. We first detail our algorithm for selecting trajectories from DROID to include in our temporal algorithm task, and then provide prompts used in the temporal ordering tasks in various settings.

\subsubsection{Temporal Filtering Algorithm for DROID}

\begin{algorithmic}
\State \textbf{Input:} Total sequences $N$, target count $K$, output directory $D$
\State \textbf{Output:} $K$ downloaded sequences with metadata

\Statex
\State Generate $K$ random indices: $\mathcal{I} = \{i_1, i_2, \ldots, i_K\}$ where $i_j \in [1, N]$

\Statex
\State Load dataset $\mathcal{D}$ from remote storage
\State Initialize metadata $\mathcal{M} = \{\}$
\State $count \gets 0$

\For{each episode $(idx, episode)$ in $\mathcal{D}$}
    \If{$idx \notin \mathcal{I}$}
        \State \textbf{continue}
        \Comment{Skip non-target indices}
    \EndIf
    
    \State $task\_desc \gets$ extract task description from $episode$
    \State $steps \gets$ list of steps in $episode$
    \State $length \gets |steps|$
    
    \If{$length = 0$ \textbf{or} $task\_desc$ is empty \textbf{or} $task\_desc =$ ``No action''}
        \State \textbf{continue}
        \Comment{Skip invalid sequences}
    \EndIf
    
    \State $seq\_num \gets count + 1$
    \State Create directory $D$/sequence\_\{$seq\_num$\}
    
    \For{each step $s$ in $steps$}
        \State Extract image from $s.\text{observation}$
        \State Save as $D$/sequence\_\{$seq\_num$\}/step\_$i$.png
    \EndFor
    
    \State $\mathcal{M}[seq\_num] \gets \{\text{original\_index}: idx,\ \text{task}: task\_desc,\ \text{length}: length\}$
    \State $count \gets count + 1$
    
    \If{$count = K$}
        \State \textbf{break}
        \Comment{All target sequences downloaded}
    \EndIf
\EndFor

\Statex
\State Save metadata $\mathcal{M}$ to $D$/download\_metadata.json
\State \Return $\mathcal{M}$

\end{algorithmic}
\noindent
\newline
\textbf{n-image Prompt for GPT models}
\begin{lstlisting}

if num_images == 2:
        order_desc = "first, second"
        example_order = "[2, 1]"
    elif num_images == 3:
        order_desc = "first, second, third"
        example_order = "[3, 1, 2]"
    elif num_images == 4:
        order_desc = "first, second, third, fourth"
        example_order = "[3, 1, 4, 2]"
    else:  # num_images == 5
        order_desc = "first, second, third, fourth, fifth"
        example_order = "[3, 1, 5, 2, 4]"
        
prompt = (
        f'Look at these {num_images} shuffled images from a robot task sequence. '
        f'The task is: "{task_description}".\n\n'
        f'The red numbers (1..{num_images}) are post-shuffle labels - they do NOT reflect chronology.\n'
        f"Your task: Determine which labeled image came {order_desc}.\n\n"
        f"Respond with ONLY a JSON array of numbers in chronological order (e.g., {example_order}).\n"
    )
\end{lstlisting}

This was used for 2,3,4, and 5 randomly shuffled images when provided to GPT models for temporal ordering.

\noindent
\newline
\textbf{n-image Prompt for non-GPT models}
\begin{lstlisting}
if num_images == 2:
        example_out = "```[2, 1]```"
    elif num_images == 3:
        example_out = "```[3, 1, 2]```"
    elif num_images == 4:
        example_out = "```[3, 1, 4, 2]```"
    else:  # num_images == 5
        example_out = "```[3, 1, 5, 2, 4]```"

    prompt = (
        f"{task_description}\n\n"
        f"These {num_images} images are shuffled. Red labels 1..{num_images} are just shuffle labels (not chronological).\n\n"
        f"Task: Determine the chronological order of these images. Which labeled image came first, second, third, etc.?\n\n"
        f"Answer as a bracketed list of labels inside backticks, earliest to latest.\n"
        f"Example: {example_out}"
    )
\end{lstlisting}

This was used for 2,3,4, and 5 randomly shuffled images when provided to non-GPT models for temporal ordering.

\noindent
\newline
\textbf{n-image Prompt for VLMs in the Segmentation Map and Bounding Box Settings}\\

When we evaluate VLMs in the bounding box or segmentation map setting, the prompt is modified to include the following line:
\begin{lstlisting}
f'Look at these {num_images} shuffled images from a robot task sequence with {viz_type_desc}. '
\end{lstlisting}
where \texttt{viz\_type\_type} could be "Bounding Box" or "Segmentation Map."

\noindent
\newline
\textbf{3/5-image Prompt for the Merge Sort Setting}
\begin{lstlisting}
prompt = (
        f'Look at these 2 images from a robot task sequence. '
        f'The task is: "{task_description}".\n\n'
        f"These images are from the same sequence but at different time points.\n"
        f"Your task: Determine which image came FIRST in the sequence.\n\n"
        f"Respond with a JSON object containing:\n"
        f'- "first_image": number of the image that came first (1 or 2)\n'
        f'- "confidence": your confidence level (0.0 to 1.0)\n'
        f'- "reasoning": brief explanation of your decision\n\n'
        f'Example: {{"first_image": 1, "confidence": 0.8, "reasoning": "The robot arm is closer to the object in image 1"}}'
    )
\end{lstlisting}

\subsection{Depth Ordering Task}

Here, the model is asked to correctly predict the ordering of objects based on their distance from the cameras. The objects in consideration are provided after filtering out non-overlapping, non-contained objects from the scene using Grounding DINO. 

\subsubsection{Depth Filtering Algorithm for DROID}

\begin{algorithmic}
\State \textbf{Input:} Total sequences $N$, target count $K$, output directory $D$, min objects $M$
\State \textbf{Output:} $K$ sequences with $\geq M$ non-overlapping objects and depth ordering

\Statex
\State Generate $K$ random indices: $\mathcal{I} = \{i_1, i_2, \ldots, i_K\}$ where $i_j \in [1, N]$

\Statex
\State Load dataset $\mathcal{D}$ from remote storage
\State Initialize accepted sequences $\mathcal{A} = \{\}$
\State $count \gets 0$

\For{each episode $(idx, episode)$ in $\mathcal{D}$}
    \If{$idx \notin \mathcal{I}$}
        \State \textbf{continue}
        \Comment{Skip non-target indices}
    \EndIf
    
    \State $steps \gets$ list of steps in $episode$
    \State $length \gets |steps|$
    
    \If{$length = 0$}
        \State \textbf{continue}
        \Comment{Skip empty sequences}
    \EndIf
    
    \Statex
    \State \textbf{Step 1: Random Image Selection}
    \State $step\_idx \gets$ random integer in $[0, length-1]$
    \State $selected\_step \gets steps[step\_idx]$
    \State Extract concatenated image $I$ from $selected\_step.\text{observation}$
    
    \Statex
    \State \textbf{Step 2: DINO Object Detection}
    \State Run DINO detector on image $I$ with prompt
    \State $\mathcal{B} \gets$ set of detected bounding boxes $\{b_1, b_2, \ldots, b_n\}$
    \State $\mathcal{L} \gets$ set of object labels $\{l_1, l_2, \ldots, l_n\}$
    
    \If{$|\mathcal{B}| < M$}
        \State \textbf{continue}
        \Comment{Insufficient objects detected}
    \EndIf
    
    \Statex
    \State \textbf{Step 3: SAM2 Segmentation}
    \State For each bounding box $b_i \in \mathcal{B}$:
    \State \quad Run SAM2 segmenter on image $I$ with box $b_i$
    \State \quad $m_i \gets$ segmentation mask for object $i$
    \State $\mathcal{M} \gets$ set of masks $\{m_1, m_2, \ldots, m_n\}$
    
    \Statex
    \State \textbf{Step 4: Marigold Depth Estimation}
    \State Run Marigold depth estimator on image $I$
    \State $D \gets$ depth map for entire image
    
    \Statex
    \State \textbf{Step 5: Depth Extraction and Validation}
    \State For each mask $m_i \in \mathcal{M}$:
    \State \quad $d_i \gets$ mean depth value in region $m_i$ from depth map $D$
    \State $\mathcal{D}_\text{obj} \gets$ set of object depths $\{d_1, d_2, \ldots, d_n\}$
    
    \State Check pairwise IoU between all masks
    \State $\mathcal{M}_\text{non-overlap} \gets$ subset of masks with IoU $< \theta$ for all pairs
    \State $num\_objects \gets |\mathcal{M}_\text{non-overlap}|$
    
    \If{$num\_objects < M$}
        \State \textbf{continue}
        \Comment{Insufficient non-overlapping objects}
    \EndIf
    
    \Statex
    \State \textbf{Step 6: Depth Ordering}
    \State Sort objects by depth: $\mathcal{O}_\text{sorted} \gets$ sort $\mathcal{M}_\text{non-overlap}$ by $d_i$ (closest to farthest)
    \State Generate depth ordering: $\text{ordering} \gets$ list of object labels in depth order
    
    \Statex
    \State \textbf{Step 7: Accept and Download}
    \State $seq\_num \gets count + 1$
    \State Create directory $D$/sequence\_\{$seq\_num$\}
    \State Save $I$ as $D$/sequence\_\{$seq\_num$\}/original\_image.jpg
    \State Save left portion of $I$ as $D$/sequence\_\{$seq\_num$\}/left\_image.jpg
    \State Save image with bounding boxes as $D$/sequence\_\{$seq\_num$\}/prompt\_image\_with\_boxes.jpg
    \State Save image with masks as $D$/sequence\_\{$seq\_num$\}/prompt\_image\_with\_masks.jpg
    \State Save pipeline results as $D$/sequence\_\{$seq\_num$\}/analysis\_results.json
    
    \State $\mathcal{A}[seq\_num] \gets \{\text{objects}: num\_objects,\ \text{ordering}: \text{ordering},\ \text{droid\_index}: idx\}$
    \State $count \gets count + 1$
    
    \If{$count = K$}
        \State \textbf{break}
        \Comment{All target sequences downloaded}
    \EndIf
\EndFor

\Statex
\State Save metadata $\mathcal{A}$ to $D$/split\_metadata.json
\State \Return $\mathcal{A}$

\end{algorithmic}

\noindent
\newline
\textbf{General VLM Prompt} \\
The general VLM prompt is as follows:
\begin{lstlisting}
    "vlm_prompt": "order the following objects from closest to farthest from camera in terms of their centroid in the image: obj1, obj2, obj3, ...",
\end{lstlisting}

\noindent
\newline
\textbf{n-object Prompt for GPT Models}\\
\begin{lstlisting}
prompt = (
        f'Look at this image from a robot task scene.\n\n'
        f'{vlm_prompt}\n\n'
        f'IMPORTANT: You must use ONLY the exact object names provided in the prompt above. '
        f'Do not invent new object names or use variations. Your response must be a JSON array. '
        f'Example: If objects provided are elephant, tiger, dragon and correct prediction of closest to farthest from camera is tiger, dragon, elephant then output the json with the order: {json.dumps(["tiger", "dragon", "elephant"])}. Do not include any other text or explanation.'
    )
\end{lstlisting}
This was used for the setting where 2,3,4, and 5 objects are present in a image provided to GPT-5 and GPT-5-Nano for depth ordering.

\noindent
\newline
\textbf{n-object Prompt for non-GPT models} \\
\begin{lstlisting}
prompt = (
        f'Look at this image from a robot task scene.\n\n'
        f'{vlm_prompt}\n\n'
        f'IMPORTANT: You must use ONLY the exact object names provided in the prompt above. '
        f'Do not invent new object names or use variations. For example '
        f'Example: If objects provided are elephant, tiger, dragon and correct prediction of closest to farthest from camera is tiger, dragon, elephant then output the string ```["tiger", "dragon", "elephant"]```'
    )
\end{lstlisting}

\noindent
\newline
\textbf{n-object Prompt for VLMs in the Segmentation Map and Bounding Box Settings} \\
When we evaluate VLMs in the bounding box or segmentation map setting, the prompt is modified to include the following
line: 
\begin{lstlisting}
f'The image shows {num_objects} objects highlighted with {visualization_description} (colored annotations).\n\n'
\end{lstlisting}
where \texttt{visualization\_description} could be "Bounding Box" or "Segmentation Map."

\noindent
\newline
\textbf{3/5-object Prompt for the Merge Sort Setting}\\
\begin{lstlisting}
prompt = (
        f'Look at this image from a robot task scene.\n\n'
        f'{vlm_prompt}\n\n'
        f'Your task: Compare the two objects "{obj1}" and "{obj2}" and determine which one is CLOSER to the camera.\n\n'
        f'IMPORTANT: You must respond with ONLY one of these exact object names: "{obj1}" or "{obj2}". '
        f'Do not use variations or invent names.\n\n'
        f'Respond with a JSON object containing:\n'
        f'- "closer_object": the name of the object that is closer (must be exactly "{obj1}" or "{obj2}")\n'
        f'- "confidence": your confidence level (0.0 to 1.0)\n'
        f'- "reasoning": brief explanation of your decision\n\n'
        f'Example: If objects provided are elephant and tiger, and elephant is closer to the camera, then output: {{"closer_object": "elephant", "confidence": 0.8, "reasoning": "Object appears larger and more prominent"}}'
    )
\end{lstlisting}

\subsection{Left-to-Right Ordering Task}

Here, the model is asked to correctly predict the left-to-right ordering of objects.

\subsubsection{Left-to-right Filtering Algorithm for DROID}
\begin{algorithmic}
\State \textbf{Input:} Image directory $I$, output directory $O$, min objects $M$
\State \textbf{Output:} Dataset of images with left-to-right ordered objects
\Statex
\State \textbf{Parameters:}
\State $\quad$ min\_gap $\gets$ minimum horizontal gap between objects (default: 0.05)
\State $\quad$ max\_iou $\gets$ maximum IoU between objects (default: 0.7)
\State $\quad$ max\_vertical\_dist $\gets$ maximum vertical distance (default: 0.5)
\State $\quad$ max\_width\_ratio $\gets$ maximum box width ratio (default: 0.7)
\\
\State Sample $N$ sequences from $I$ 
\For{each sequence $S$ in sampled sequences}
    \State \textbf{Step 1: Image Selection}
    \State $img \gets$ random image from $S$
    \State Load image dimensions: $(W, H) \gets$ size of $img$
    \State Extract left portion: $img \gets img[:, :W \times 0.33]$
    \Statex
    \State \textbf{Step 2: Object Detection}
    \State Stage 1: Query GPT-5 with $img$ to identify object categories
    \State Stage 2: Run Grounding DINO with GPT-5 labels as text prompts
    \State Extract results: $\mathcal{L} \gets$ labels, $\mathcal{B} \gets$ bboxes, $\mathcal{S} \gets$ scores
    \If{$|\mathcal{L}| < M$}
        \State \textbf{continue}
        \Comment{Insufficient detections}
    \EndIf
    \Statex
    \State \textbf{Step 3: Scene Object Filtering}
    \State $\mathcal{L}_{ns} = \{\}$, $\mathcal{B}_{ns} = \{\}$, $\mathcal{S}_{ns} = \{\}$
    \State $count_{scene} \gets 0$
    \For{each object $i \in \{1, \ldots, |\mathcal{L}|\}$}
        \State Compute box dimensions: $w_i \gets \mathcal{B}_i[2] - \mathcal{B}_i[0]$, $h_i \gets \mathcal{B}_i[3] - \mathcal{B}_i[1]$
        \State Compute width ratio: $r_w \gets w_i / W$
        
        \If{$r_w > \text{max\_width\_ratio}$}
            \State $count_{scene} \gets count_{scene} + 1$
            \State \textbf{continue}
            \Comment{Filter large scene objects}
        \EndIf
        
        \State $\mathcal{L}_{ns} \gets \mathcal{L}_{ns} \cup \{\mathcal{L}_i\}$
        \State $\mathcal{B}_{ns} \gets \mathcal{B}_{ns} \cup \{\mathcal{B}_i\}$
        \State $\mathcal{S}_{ns} \gets \mathcal{S}_{ns} \cup \{\mathcal{S}_i\}$
    \EndFor
    
    \If{$|\mathcal{L}_{ns}| < M$}
        \State \textbf{continue}
        \Comment{Insufficient objects after scene filtering}
    \EndIf
    \Statex
    \State \textbf{Step 4: Clear Ordering Filter}
    \State \textbf{Step 4a: Sort by left edge}
    \State Sort by x-coordinate: $idx_{sorted} \gets$ argsort($\mathcal{B}_{ns}[:, 0]$)
    \State $\mathcal{L}_{sorted} \gets [\mathcal{L}_{ns}[i] : i \in idx_{sorted}]$
    \State $\mathcal{B}_{sorted} \gets [\mathcal{B}_{ns}[i] : i \in idx_{sorted}]$
    \State $\mathcal{S}_{sorted} \gets [\mathcal{S}_{ns}[i] : i \in idx_{sorted}]$
    \State \textbf{Step 4b: Vertical clustering}
    \State Compute y-centers: $y_c[i] \gets (\mathcal{B}_{sorted}[i][1] + \mathcal{B}_{sorted}[i][3]) / 2$ for all $i$
    \State $best\_cluster \gets \emptyset$
    \State $best\_size \gets 0$
     \For{each object $i \in \{1, \ldots, |\mathcal{L}_{sorted}|\}$}
        \State $cluster_i \gets \emptyset$
        \For{each object $j \in \{1, \ldots, |\mathcal{L}_{sorted}|\}$}
            \State $y_{diff} \gets |y_c[j] - y_c[i]|$
            \If{$y_{diff} / H \leq \text{max\_vertical\_dist}$}
                \State $cluster_i \gets cluster_i \cup \{j\}$
            \EndIf
        \EndFor
        
        \If{$|cluster_i| > best\_size$}
            \State $best\_cluster \gets cluster_i$
            \State $best\_size \gets |cluster_i|$
        \EndIf
    \EndFor
    \If{$|best\_cluster| < M$}
        \Comment{Insufficient vertically-aligned objects}
    \EndIf
    \State $\mathcal{L}_{clust} \gets [\mathcal{L}_{sorted}[i] : i \in best\_cluster]$
    \State $\mathcal{B}_{clust} \gets [\mathcal{B}_{sorted}[i] : i \in best\_cluster]$
    \State $\mathcal{S}_{clust} \gets [\mathcal{S}_{sorted}[i] : i \in best\_cluster]$
    \State \textbf{Step 4c: Remove highly overlapping boxes}
    \State $to\_keep \gets []$
    \Comment{List of indices to keep (maintains insertion order)}
    
    \For{each object $i \in \{0, \ldots, |\mathcal{L}_{clust}| - 1\}$}
        \State $keep \gets$ True
        \For{each index $j \in to\_keep$}
            \State $iou_{ij} \gets$ IoU($\mathcal{B}_{clust}[i]$, $\mathcal{B}_{clust}[j]$)
            \If{$iou_{ij} > \text{max\_iou}$}
                \Comment{High overlap detected}
                \If{$\mathcal{S}_{clust}[i] > \mathcal{S}_{clust}[j]$}
                    \State Remove $j$ from $to\_keep$
                    \Comment{Current has higher confidence}
                \Else
                    \State $keep \gets$ False
                    \Comment{Existing has higher confidence}
                    \State \textbf{break}
                    \Comment{Skip current object}
                \EndIf
            \EndIf
        \EndFor
        
        \If{$keep$}
            \State Append $i$ to $to\_keep$
        \EndIf
    \EndFor
    \State $\mathcal{L}_{clear} \gets [\mathcal{L}_{clust}[i] : i \in to\_keep]$
    \State $\mathcal{B}_{clear} \gets [\mathcal{B}_{clust}[i] : i \in to\_keep]$
    \State $\mathcal{S}_{clear} \gets [\mathcal{S}_{clust}[i] : i \in to\_keep]$
    \If{$|\mathcal{L}_{clear}| < M$}
        \State \textbf{continue}
        \Comment{Insufficient unambiguous orderings}
    \EndIf
    \State \textbf{Step 5: Remove Duplicate Labels}
    \State $keep\_idx \gets \{i : |\{j : \mathcal{L}_{clear}[j] = \mathcal{L}_{clear}[i]\}| = 1\}$
    \State $\mathcal{L}_{final} \gets [\mathcal{L}_{clear}[i] : i \in keep\_idx]$
    \State $\mathcal{B}_{final} \gets [\mathcal{B}_{clear}[i] : i \in keep\_idx]$
    \State $\mathcal{S}_{final} \gets [\mathcal{S}_{clear}[i] : i \in keep\_idx]$
    
    \If{$|\mathcal{L}_{final}| < M$}
        \State \textbf{continue}
        \Comment{Insufficient objects or ambiguous ordering}
    \EndIf
    
    \State Save to $O_S$
\EndFor
\end{algorithmic}

\noindent
\newline
\textbf{General VLM Prompt} \\
The general VLM prompt is as follows:
\begin{lstlisting}
    "vlm_prompt": "Arrange the following objects in order from leftmost to rightmost in the image: obj1, obj2, obj3, ..."
\end{lstlisting}

\noindent
\newline
\textbf{n-object Prompt for all VLMs} \\
\begin{description}
  \item[\texttt{selected\_objects}] A list of $n$ randomly shuffled detected object labels,
        e.g., \texttt{["table", "chair", "door"]}.
\end{description}
\begin{lstlisting}
prompt = (
    "Arrange the following objects in order from leftmost to rightmost in the image: "
    + ", ".join(selected_objects)
    + ".\nReturn only the ordered list in JSON format: "
    + '{"ordered_objects": ["obj1", "obj2", ...]}'
)
\end{lstlisting}
These prompts were used for cases with 2, 3, 4, and 5 objects present in an image (or an image overlaid with segmentation maps or bounding boxes), where the VLMs were expected to correctly predict their left-to-right order.

\noindent
\newline
\textbf{3/5-object Prompt for the Merge Sort Setting}\\
\begin{lstlisting}
    prompt = (
        f"Which object is to the LEFT of the other in the image: '{obj1}' or '{obj2}'?\n"
        f"Return ONLY the name of the leftmost object in JSON format: "
        f'{{"leftmost_object": "obj_name"}}'
    )
\end{lstlisting}

\section{Reinforcement Learning Experiments}

This section collates the reinforcement learning (RL) details of our experiments: prompts, the final output token limit, and training curves.

\subsection{Overview of Experiments}

We ran six GRPO RL experiments on 2B models described in Table \ref{tab:reinforcement_learning}:
\begin{itemize}[leftmargin=1.25em]
    \item Two runs for \textbf{Temporal Ordering} (3-image and 5-image settings)
    \item Two runs for \textbf{Depth Ordering} (3-object and 5-object settings)
    \item Two runs for \textbf{Left-to-Right Ordering} (3-object and 5-object settings)
\end{itemize}

\subsection{Training Configuration}

Here we summarise the key configuration choices made during RL-finetuning.

\paragraph{Reward functions.}  
We experimented with a variety of reward signal formulations:  
\begin{itemize}  
  \item A fully sparse reward (e.g., +1 for correct ordering, 0 otherwise)  
  \item A rank-based score using Spearman’s rank correlation (i.e., penalising the deviation of the predicted ordering from the ground truth in Spearman’s rho)  
  \item Ultimately we settled on a Kendall $\tau$ based reward (percentage of correct pairwise orderings)
\end{itemize}

\paragraph{RL algorithm and scale.}  
We used the off-the-shelf GRPO algorithm (Group Relative Policy Optimization)  
In our setup:  
\begin{itemize}  
  \item We trained on 700 unique prompt examples.  
  \item We conducted training for 1,000 gradient/update steps $\sim$ 1.43 epochs.  
  \item Per prompt, we sampled 6 generations at each step ( group size = 6).  
\end{itemize}

\subsection{Global Settings}

\textbf{Max output tokens:} We set the maximum output token limit to \textbf{700 tokens} across all 2B RL tasks. %

\medskip
\noindent\textbf{Reasoning first:} In all tasks, we nudge the model to write a short English reasoning trace before clearly stating the final ordering. That lets us treat the text as a trajectory for optimization and helps the model overcome any default ordering bias.

\subsection{Temporal Ordering Task (with stamping + randomized example output)}

Here the model sees a shuffled sequence of \emph{multiple} images (either 3 or 5) from a robot task sequence and must decide which came first, second, etc.

\paragraph{Key implementation details (temporal only).}
\begin{itemize}[leftmargin=1.25em]
    \item \textbf{Stamped labels:} We overlay small red numeric markers on each image so the model can unambiguously refer to them (1, 2, 3, \ldots). This stamping is \emph{only used} for the temporal task.
    \item \textbf{Randomized example output:} Each training example includes a \emph{randomized} final-answer example like \texttt{[3, 1, 2]}. The randomization is deterministic per example ID and helps avoid mode collapse onto a single example format. This randomization is \emph{only used} for the temporal task.
\end{itemize}

\noindent
\textbf{3-image version (system prompt skeleton).}

\begin{lstlisting}
Look at these 3 shuffled images from a robot task sequence.
The robot task is: "<task description>".

The red numbers (1, 2, 3) are post-shuffle labels -
they do NOT reflect chronology.
Your task: Reason about which labeled image came first,
second, third.

Feel free to write your process/reasoning in English
before outputting your final answer.
Example final answer (for formatting only):
if the correct order is third, first, second,
then at the end output [3, 1, 2]
\end{lstlisting}

The example final answer is randomized throughout training.
\newline
\\
\textbf{5-image version (system prompt skeleton).}
Same as above, but for five images and with labels \texttt{1..5}.

\subsection{Depth Ordering Task}

For depth, the model sees a \emph{single} image from the robot workspace and must order objects from closest to farthest .
\newline
\\
\textbf{Prompt.}
\begin{lstlisting}
You are a visual reasoner. You will be shown a single image
from a robot workspace. You may write notes or reason before
answering, but make sure to clearly state the final ordering.

Order the following objects from closest to farthest from camera
in terms of their centroid in the image:
robot arm, curtain, window, blanket, pillow, table
\end{lstlisting}

\subsection{Left-to-Right Ordering Task}

Similarly, for left-to-right we use one image and ask the model to order objects horizontally from left to right.
\newline
\\
\textbf{Prompt.}
\begin{lstlisting}
You are a visual reasoner. You will be shown a single image
from a robot workspace. You may write notes or reason before
answering, but make sure to clearly state the final ordering.
\end{lstlisting}

\subsection{Training Curves}

\begin{figure}[H]
    \centering
    \includegraphics[width=0.8\linewidth]{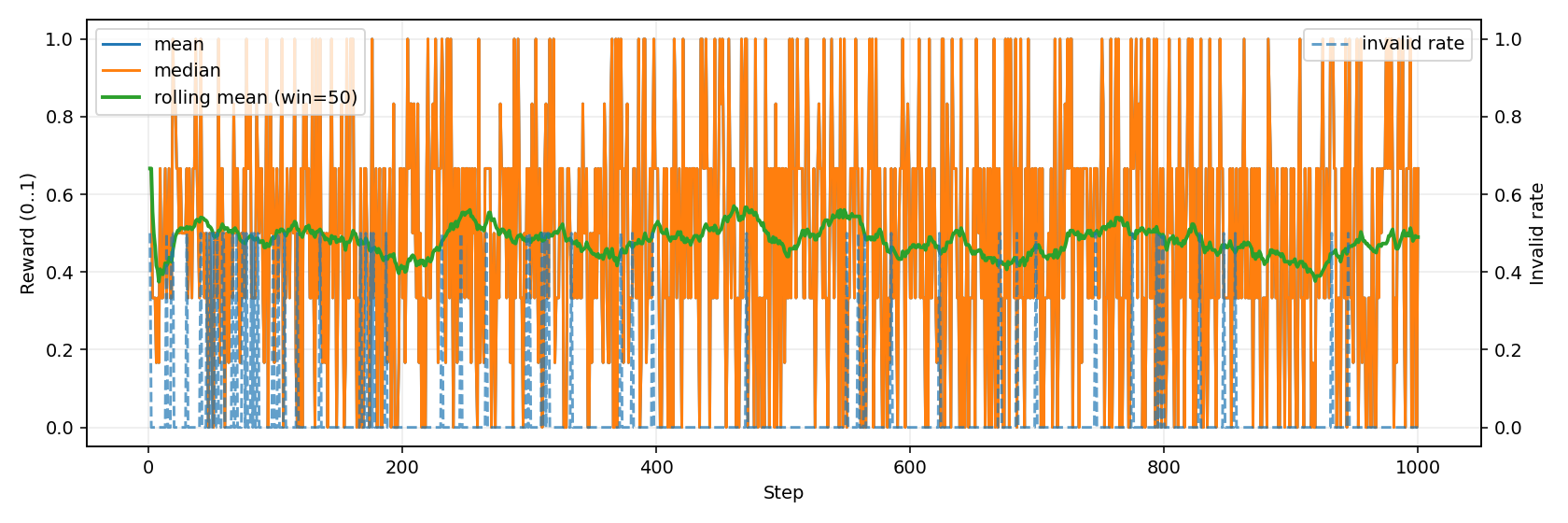}
    \caption{Temporal task, 3 img}
    \label{fig:reward_curve_temporal_3}
\end{figure}

\begin{figure}[H]
    \centering
    \includegraphics[width=0.8\linewidth]{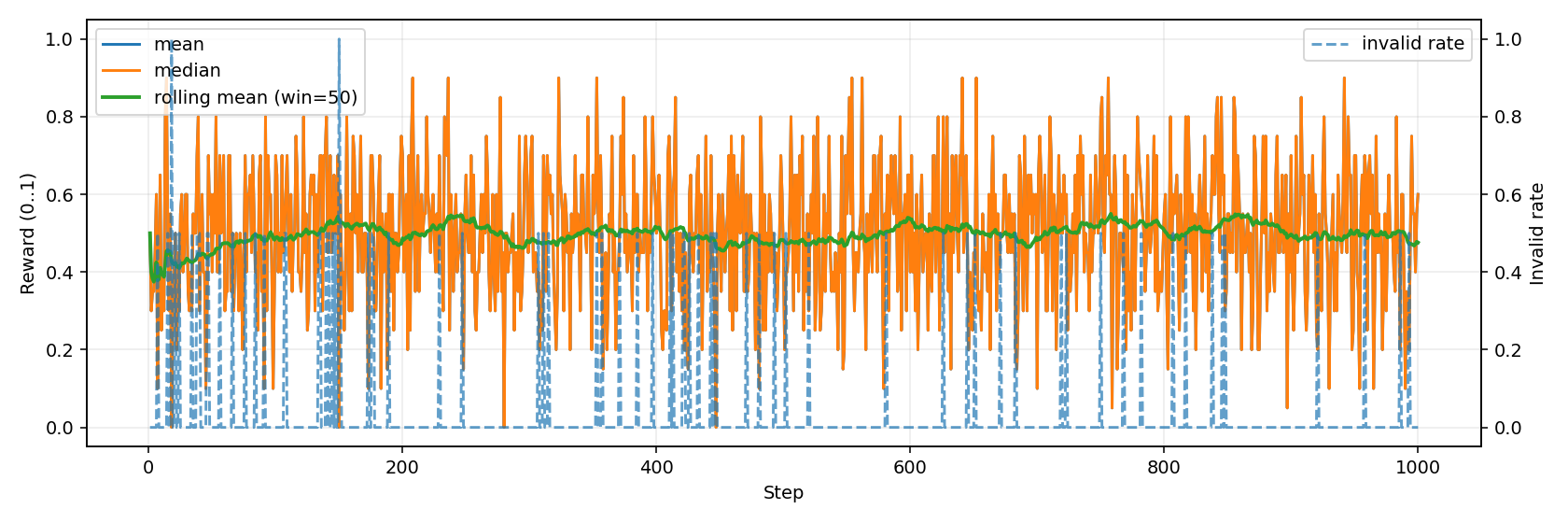}
    \caption{Temporal task, 5 image}
    \label{fig:reward_curve_temporal_5}
\end{figure}

\begin{figure}[H]
    \centering
    \includegraphics[width=0.8\linewidth]{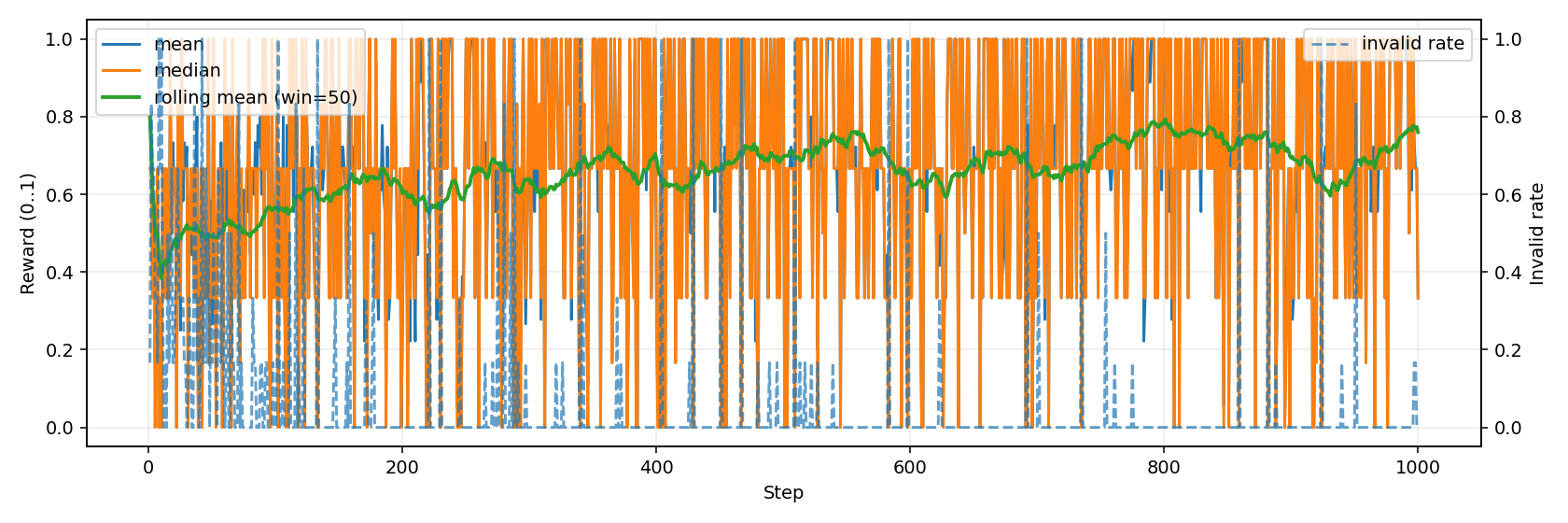}
    \caption{Depth task,3 Objects}
    \label{fig:reward_curve_depth_3}
\end{figure}

\begin{figure}[H]
    \centering
    \includegraphics[width=0.8\linewidth]{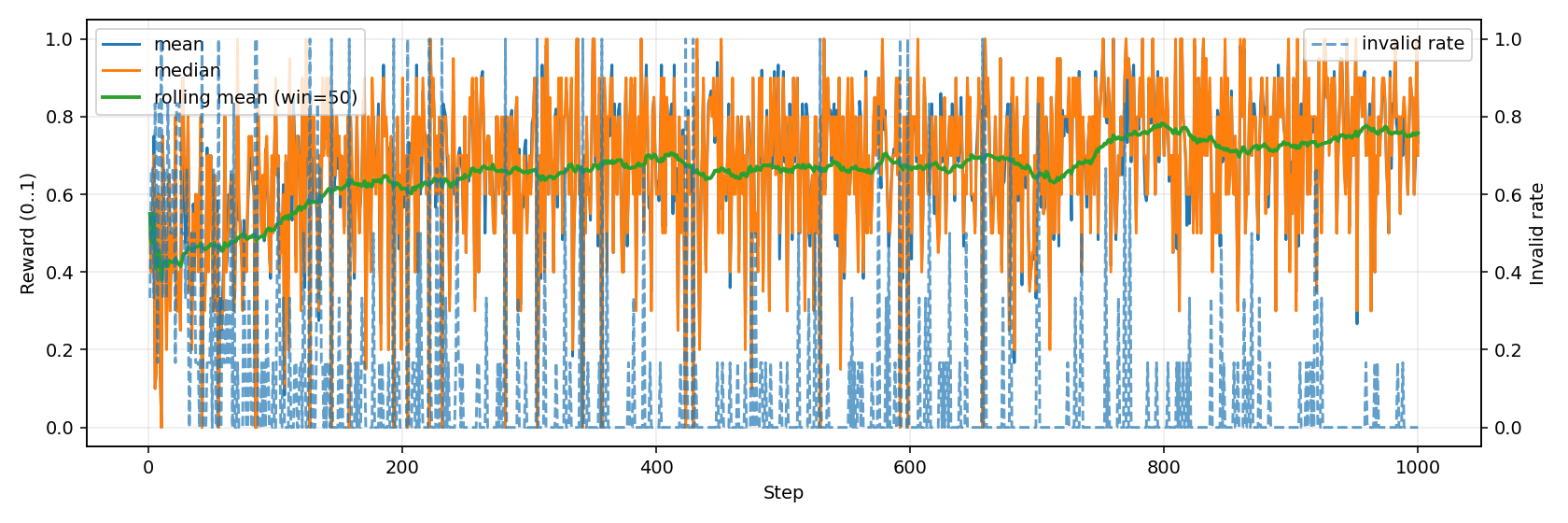}
    \caption{Depth task, 5 objects}
    \label{fig:reward_curve_depth_5}
\end{figure}

\begin{figure}[H]
    \centering
    \includegraphics[width=0.8\linewidth]{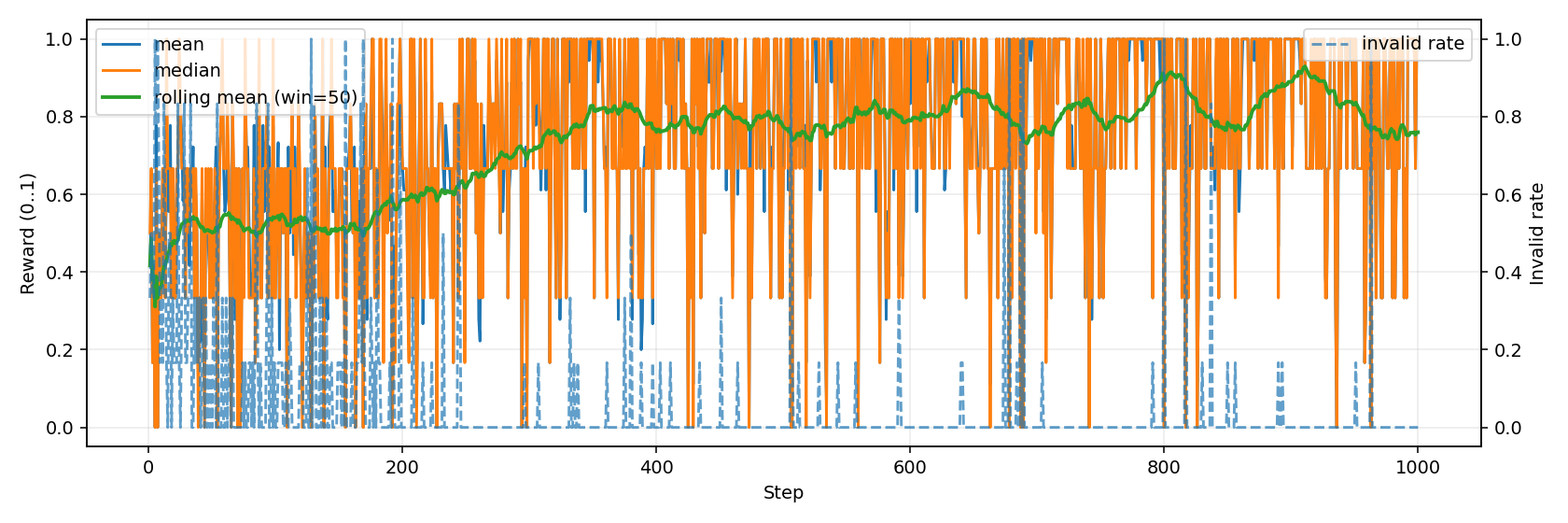}
    \caption{Left-to-Right task, 3 Objects}
    \label{fig:reward_curve_l2r_3}
\end{figure}

\begin{figure}[H]
    \centering
    \includegraphics[width=0.8\linewidth]{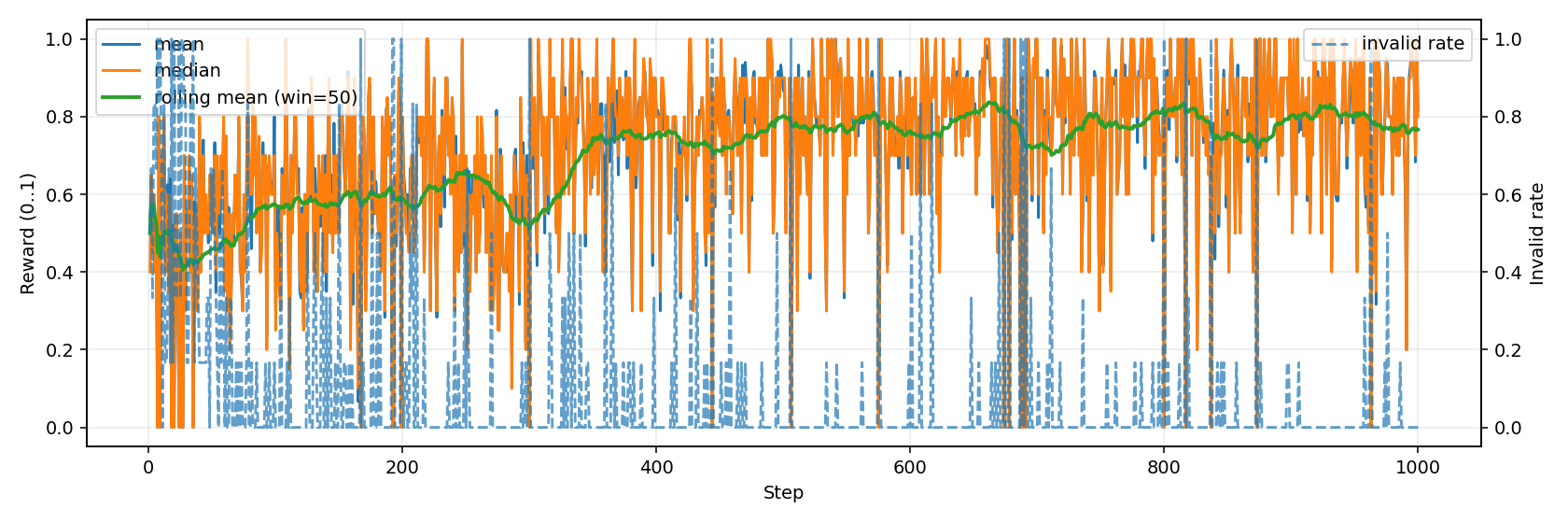}
    \caption{Left-to-Right task, 5 Objects}
    \label{fig:reward_curve_l2r_5}
\end{figure}

Figures \ref{fig:reward_curve_temporal_3}-\ref{fig:reward_curve_l2r_5}
show the reward curves for all six GRPO runs (temporal, depth, left-to-right; 3-object/image and 5-object/image) for Qwen2-VL-2B-Instruct.

\medskip
\noindent\textit{Notes.}
\begin{itemize}[leftmargin=1.25em]
    \item Stamping and randomized example outputs are \textbf{temporal-only}. Depth and left-to-right use a single image, no stamping, no randomized example.
    \item All RL runs used a max generation budget of \textbf{700 tokens}.
    \item The reward on the y-axis is the kendall reward (fraction of correct pairwise orderings in model output)
\end{itemize}

\newpage 
\onecolumn
\section{Task Examples}
In this section, we provide examples of temporal, depth and spatial ordering tasks in the base as well as bounding box and segmentation map settings.

\subsection{Examples of Temporal Ordering Tasks}

\begin{figure}[H]
    \centering
    \includegraphics[width=\textwidth]{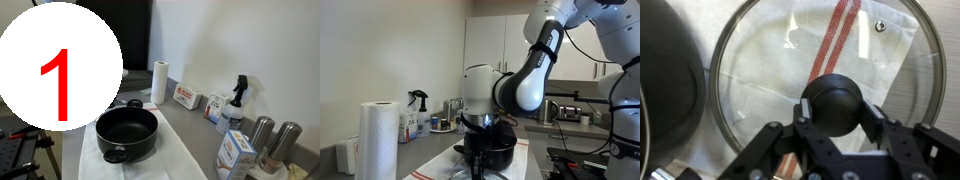} \\
    \includegraphics[width=\textwidth]{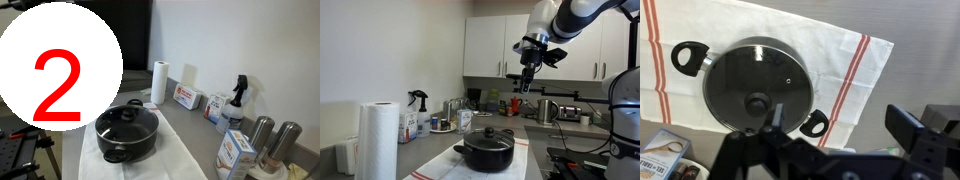} \\
    \includegraphics[width=\textwidth]{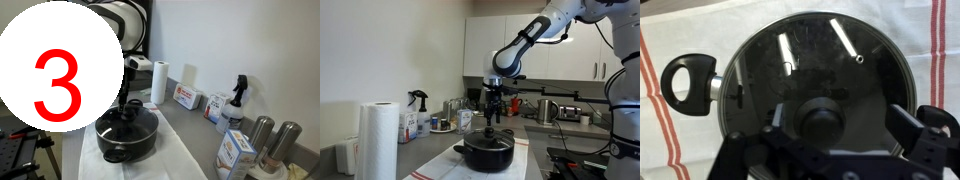} \\
    \includegraphics[width=\textwidth]{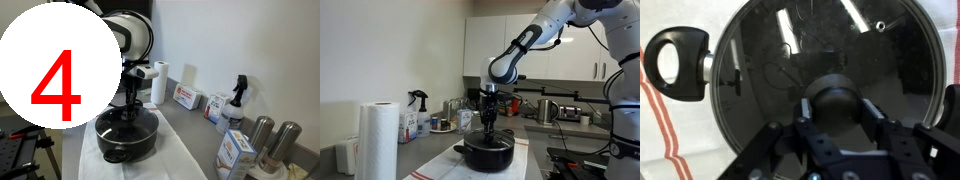} \\
    \includegraphics[width=\textwidth]{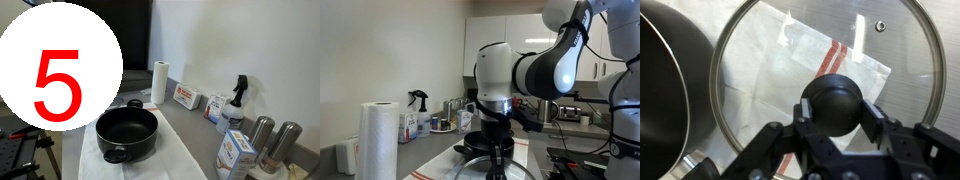}
    \caption{Task Description: ``Remove the lid from the pot | Remove the lid from the black pot. | Remove the glass lid from the pot." Correct Order: 2,3,4,5,1}
    \label{fig:temporal_base_1}
\end{figure}

\begin{figure*}[htbp]
    \centering
    \includegraphics[width=\textwidth]{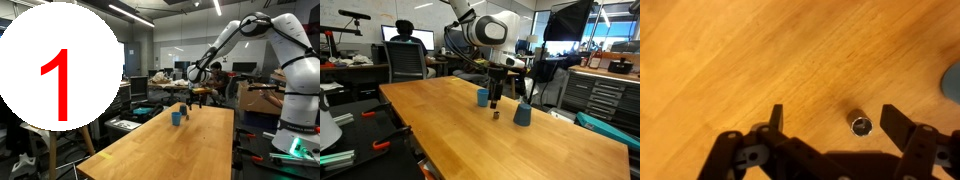} \\
    \includegraphics[width=\textwidth]{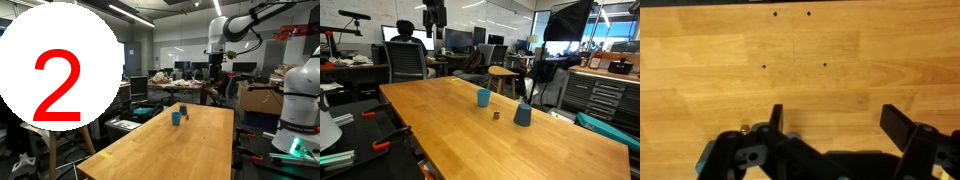} \\
    \includegraphics[width=\textwidth]{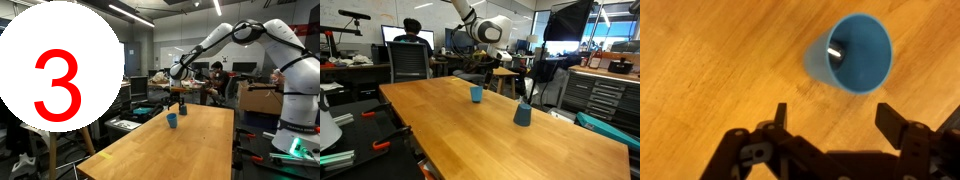} \\
    \includegraphics[width=\textwidth]{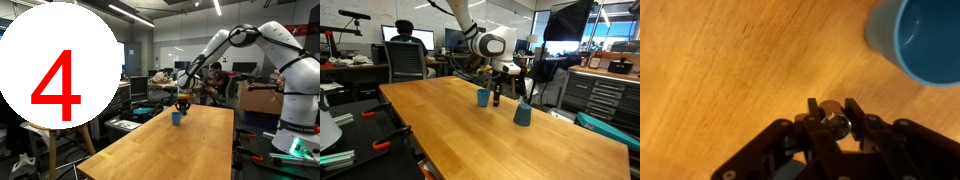} \\
    \includegraphics[width=\textwidth]{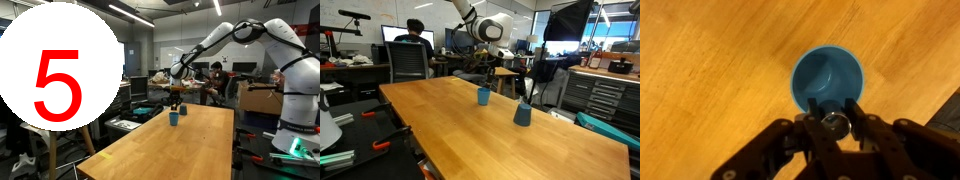}
    \caption{Task Description: ``Pick up the object and put it in the light blue cup | Place the metal object inside the light blue cup | Put the silver object in the light blue cup" \\ Correct Order: 2,1,4,5,3}
    \label{fig:temporal_base_2}
\end{figure*}

\begin{figure*}[htbp]
    \centering
    \includegraphics[width=\textwidth]{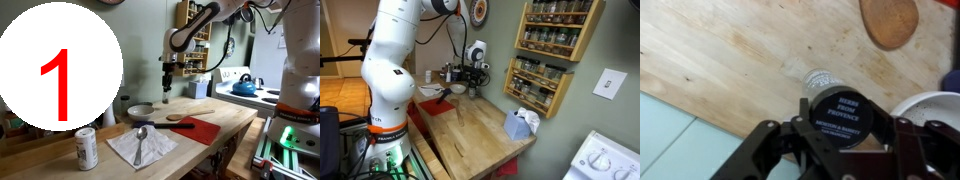} \\
    \includegraphics[width=\textwidth]{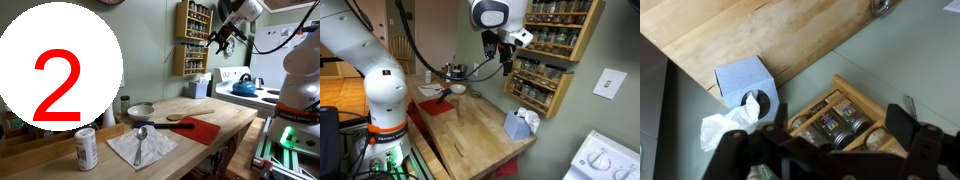} \\
    \includegraphics[width=\textwidth]{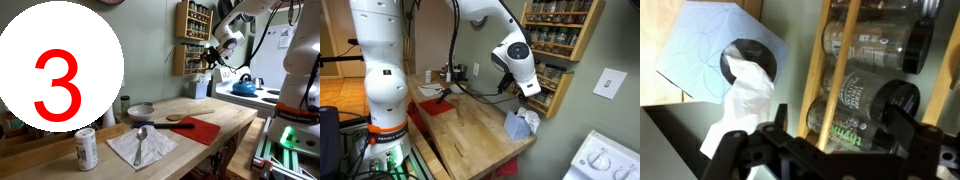} \\
    \includegraphics[width=\textwidth]{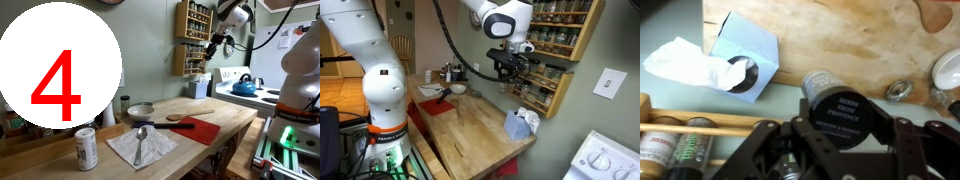} \\
    \includegraphics[width=\textwidth]{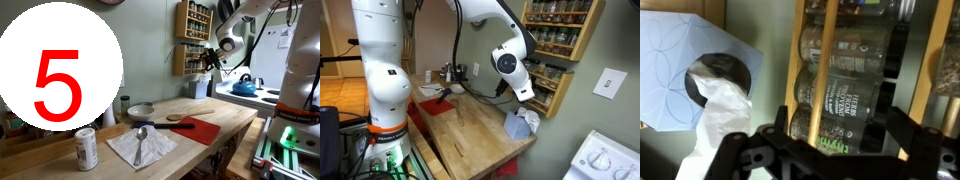}
    \caption{Task Description: ``Pick up the third spice bottle from the right on the bottom shelf of the bottom spice rack and put it on the table | Remove the third container from the right on the bottom rack and put it on the table | Put the spice bottle on the counter" \\ Correct Order: 2,5,3,4,1}
    \label{fig:temporal_base_3}
\end{figure*}

\begin{figure}[H]
    \centering
    \includegraphics[width=\textwidth]{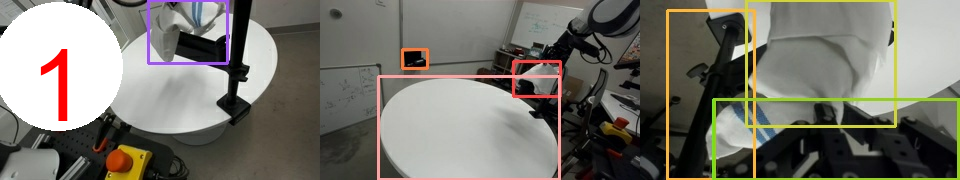} \\
    \includegraphics[width=\textwidth]{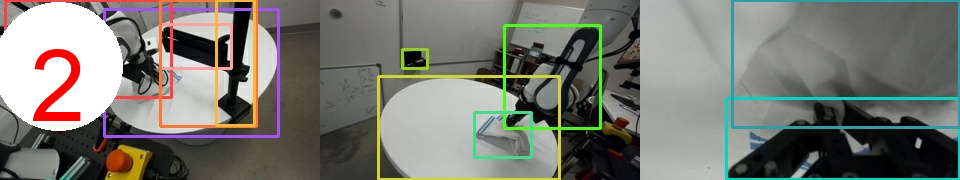} \\
    \includegraphics[width=\textwidth]{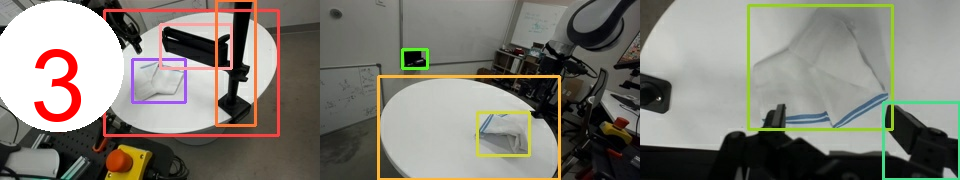} \\
    \caption{Task Description: ``Hang the towel on the black stand | Put the towel on the black rod | Pick up the cloth from the table and place it on the black object" \\ Correct Order: 3, 2, 1}
    \label{fig:temporal_bbox_1}
\end{figure}

\begin{figure}[H]
    \centering
    \includegraphics[width=\textwidth]{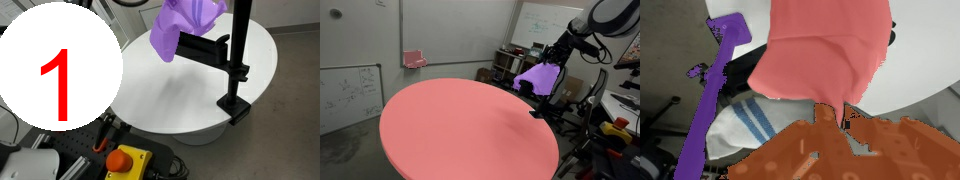} \\
    \includegraphics[width=\textwidth]{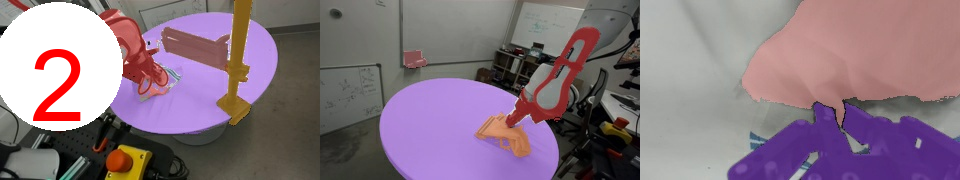} \\
    \includegraphics[width=\textwidth]{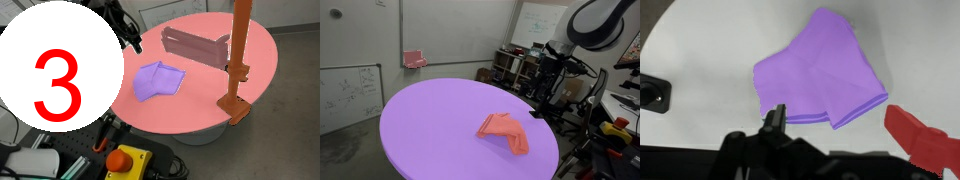} \\
    \caption{Task Description: ``Hang the towel on the black stand | Put the towel on the black rod | Pick up the cloth from the table and place it on the black object" \\ Correct Order: 3, 2, 1}
    \label{fig:temporal_segmap_1}
\end{figure}

We show examples of temporal ordering tasks across three settings: base (Figures \ref{fig:temporal_base_1}-\ref{fig:temporal_base_3}), bounding box (Figure \ref{fig:temporal_bbox_1}), and segmentation map (Figure \ref{fig:temporal_segmap_1}).\\
\newline
\noindent
\textbf{Base Setting Tasks}

\clearpage
\pagebreak[4]
\noindent
\textbf{Bounding Box Setting Task}

\noindent
\textbf{Segmentation Map Setting Task}

\clearpage
\newpage
\twocolumn
\subsection{Examples of Depth Ordering Tasks}

\begin{figure}[H]
    \centering
    \includegraphics[width=\linewidth]{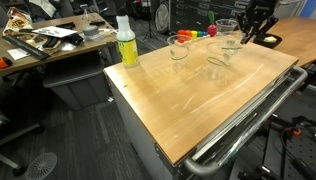} \\
    \caption{``order the following objects from closest to farthest from camera in terms of their centroid in the image: bin, robot arm, bowl, bottle, bowl cup" \\ Correct Order: "robot arm", "bottle", "bowl cup", "bowl", "bin"}
    \label{fig:depth_base_1}
\end{figure}

\begin{figure}[H]
    \centering
    \includegraphics[width=\linewidth]{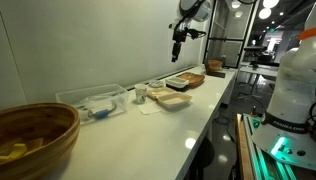} \\
    \caption{``order the following objects from closest to farthest from camera in terms of their centroid in the image: container, cup, bottle, tray, robot arm" \\ Correct Order: "container", "cup", "tray", "robot arm", "bottle"}
    \label{fig:depth_base_2}
\end{figure}

\begin{figure}[H]
    \centering
    \includegraphics[width=\linewidth]{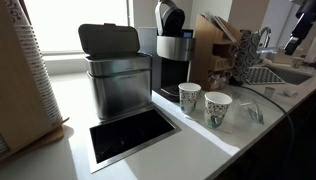} \\
    \caption{``order the following objects from closest to farthest from camera in terms of their centroid in the image: sink, faucet, trash bin, cardboard box packet, coffee maker" \\ Correct Order: "cardboard box packet", "trash bin", "coffee maker", "sink", "faucet"}
    \label{fig:depth_base_3}
\end{figure}

\begin{figure}[H]
    \centering
    \includegraphics[width=\linewidth]{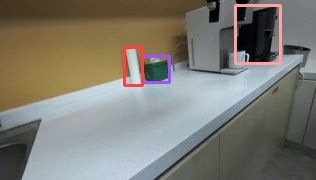} \\
    \caption{``order the following objects from closest to farthest from camera in terms of their centroid in the image: napkin dispenser, paper towel roll, coffee machine" \\ Correct Order: "paper towel roll", "napkin dispenser", "coffee machine"}
    \label{fig:depth_bbox_1}
\end{figure}

\begin{figure}[H]
    \centering
    \includegraphics[width=\linewidth]{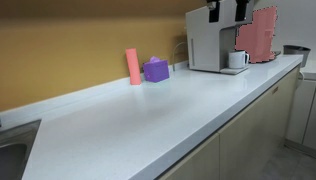} \\
    \caption{``order the following objects from closest to farthest from camera in terms of their centroid in the image: napkin dispenser, paper towel roll, coffee machine" \\ Correct Order: "paper towel roll", "napkin dispenser", "coffee machine"}
    \label{fig:depth_segmap_1}
\end{figure}

We show examples of depth ordering tasks across three settings: base (Figures \ref{fig:depth_base_1}-\ref{fig:depth_base_3}), bounding box (Figure \ref{fig:depth_bbox_1}), and segmentation map (Figure \ref{fig:depth_segmap_1}) .

\subsubsection{Base Setting Task}

\noindent
\textbf{Bounding Box Setting Task}

\textbf{Segmentation Map Setting Task}

\clearpage
\subsection{Examples of Left-to-Right Tasks}

\begin{figure}[H]
    \centering
    \includegraphics[width=\linewidth]{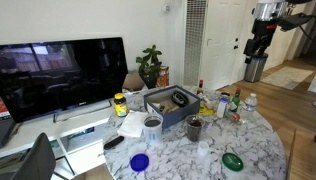} \\
    \caption{``Arrange the following objects in order from leftmost to rightmost in the image: rug, chair, plant, table, door" \\ Correct Order: "chair", "table", "plant", "door", "rug"}
    \label{fig:ltr_base_1}
\end{figure}

\begin{figure}[H]
    \centering
    \includegraphics[width=\linewidth]{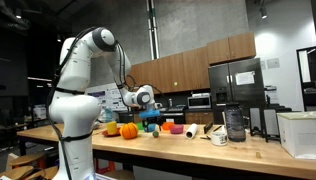} \\
    \caption{``Arrange the following objects in order from leftmost to rightmost in the image: robot, paper towel, box, microwave, pumpkin" \\ Correct Order: "robot", "pumpkin", "paper towel", "microwave", "box"}
    \label{fig:ltr_base_2}
\end{figure}

\begin{figure}[H]
    \centering
    \includegraphics[width=\linewidth]{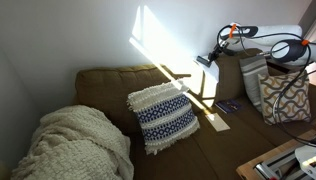} \\
    \caption{``Arrange the following objects in order from leftmost to rightmost in the image: blanket, lamp, pillow blanket, table, book" \\ Correct Order: "blanket", "pillow blanket", "lamp", "book", "table"}
    \label{fig:ltr_base_3}
\end{figure}

\begin{figure}[H]
    \centering
    \includegraphics[width=\linewidth]{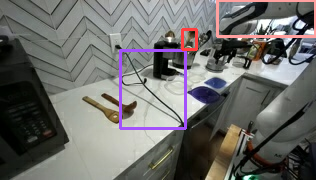} \\
    \caption{``Arrange the following objects in order from leftmost to rightmost in the image: cable, robot arm, spoon" \\ Correct Order: "cable", "spoon", "robot arm"}
    \label{fig:ltr_bbox_1}
\end{figure}

\begin{figure}[H]
    \centering
    \includegraphics[width=\linewidth]{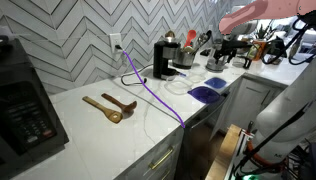} \\
    \caption{``Arrange the following objects in order from leftmost to rightmost in the image: cable, robot arm, spoon" \\ Correct Order: "cable", "spoon", "robot arm"}
    \label{fig:ltr_segmap_1}
\end{figure}

We show examples of left-to-right ordering tasks across three settings: base (Figures \ref{fig:ltr_base_1}-\ref{fig:ltr_base_3}), bounding box (Figure \ref{fig:ltr_bbox_1}), and segmentation map (Figure \ref{fig:ltr_segmap_1}).\\

\noindent
\textbf{Base Setting Task}

\textbf{Bounding Box Setting Task}

\textbf{Segmentation Map Setting Task}

\section{LLM Usage}
We used LLM-generated code (which was reviewed by the authors) for implementing some experiments.

\section{Token Limits for Reasoning Levels}
\begin{table}[H]
\tiny
\centering
\begin{tabular}{c|ccc|ccc|ccc}
\toprule
\multirow{2}{*}{\# Images/Objects} & \multicolumn{3}{c}{Temporal Ordering Task} & \multicolumn{3}{c}{Depth Ordering Task} & \multicolumn{3}{c}{Left-to-Right Ordering Task}
\\
& Low & Medium & High & Low & Medium & High & Low & Medium & High \\
\midrule
2 & 419 & 1204 & 2912 & 721 & 1487 & 5839 & 960 & 1664 & 5376 \\
3 & 535 & 1619 & 3859 & 788 & 2067 & 5778 & 896 & 2432 & 4416 \\
4 & 668 & 2062 & 5012 & 985 & 2262 & 6745 & 1024 & 2496 & 4992\\
5 & 801 & 2496 & 121{,}436 & 1052 & 2781 & 7194 & 1216 & 3136 & 6784 \\
Mergesort 3 & N/A & N/A & 5701 & N/A & N/A & 2653 & N/A & N/A & 6400 \\
Mergesort 5 & N/A & N/A & 6036 & N/A & N/A & 2800 & N/A & N/A & 11264 \\
\bottomrule
\end{tabular}
\caption{Maximum output tokens permitted across reasoning levels and number of ordering elements.}
\label{tab:reasoning_levels}
\end{table}

\section{Downstream Reasoning Performance on Qwen3-VL-8B-Thinking}
\label{app:thinking-results}

We also evaluated Qwen3-VL-8B-Thinking on our spatiotemporal ordering and downstream reasoning tasks after finetuning the model on pipeline-generated data. However, we find that Qwen3-VL-8B-Instruct checkpoint outperforms the Thinking model on a similar token budget across our downstream reasoning tasks, which is why we report our Qwen3-VL-8B-Instruct results in Section \ref{sec:experiments}. We attribute this gap in performance to Qwen3-VL-8B-Thinking's behavior of expending tokens on reasoning, which often causes it to exhaust the token budget prior to generating a valid response. We provide our Qwen3-VL-8B-Thinking results in Table \ref{tab:8b-thinking-results} for completeness.

\begin{table}[H]
\centering
\setlength{\tabcolsep}{4pt}
\resizebox{\textwidth}{!}{%
\begin{tabular}{c|ccc|ccccc}
\toprule
Model & \multicolumn{3}{c|}{3 Image/Object Spatiotemporal Ordering} & \multicolumn{5}{c}{Downstream Reasoning} \\
& Temporal & Depth & Left-to-Right & SAT & RoboFAC (MCQ) & RoboFAC (Open-Ended) & Value Est. (RT-1) & Value Est. (Dobb-E) \\
\midrule
Base & 18.33 & 29.00 & 45.00 & 48.00 & 38.54 & 00.34 & 0.0000 & 0.0005\\
SFT (Temporal) & \textbf{44.00} & 25.33 & 44.00 & 52.00 & 37.55 & 00.25 & 0.0000 & 0.0024 \\
SFT (Depth) & 23.67 & 35.00 & 49.00 & 42.67 & 33.91 & 00.29 & 0.0002 & 0.0022 \\
SFT (Left-to-Right) & 20.33 & 24.33 & \textbf{66.33} & 52.00 & 38.91 & 00.30 & \textbf{0.0183} & \textbf{0.0058}\\
GRPO (Temporal) & 16.33 & 27.67 & 45.33 &  45.33 & 21.51 & 00.03 & 0.0000 & 0.0010\\
GRPO (Depth) & 22.67 & \textbf{52.33} & 29.00 & 44.67 & 13.59 & 00.00 & 0.0081 & 0.0036 \\
GRPO (Left-to-Right) & 21.00 & 15.33 & 44.67 & \textbf{59.33} & \textbf{46.09} & \textbf{00.74} & 0.0016 & 0.0002\\
\bottomrule
\end{tabular}%
}
\caption{Finetuned Qwen3VL-8B-Thinking Accuracies on Spatiotemporal Ordering and Downstream Reasoning Tasks}
\label{tab:8b-thinking-results}
\end{table}

\end{document}